\documentclass[10pt, fleqn]{article}
\usepackage[utf8]{inputenc}
\DeclareUnicodeCharacter{2212}{$-$}
\usepackage{graphicx}
\usepackage[fleqn]{amsmath}
\usepackage[version=4]{mhchem}
\usepackage{siunitx}
\usepackage{longtable,tabularx}
\usepackage{nccmath}
\usepackage{amsmath}
\usepackage[superscript]{cite}
\usepackage[margin=1in]{geometry}
\usepackage[toc,page]{appendix}
\usepackage{setspace}
\usepackage{amsfonts, gensymb, bm, times}
\usepackage{tabu}
\usepackage{listings}
\usepackage{afterpage}
\usepackage{titlesec}
\usepackage[labelfont=bf]{caption}
\usepackage{fancyhdr}
\usepackage{pgf}
\usepackage{float}
\usepackage{subcaption}
\usepackage{cleveref}
\usepackage{textcomp}
\usepackage{pict2e}
\usepackage{wasysym}
\usepackage{multicol}
\usepackage{lipsum}
\usepackage{adjustbox}
\usepackage{indentfirst}
\usepackage{enumitem}
\usepackage[outdir=./]{epstopdf}
\newenvironment{Figure}
  {\par\medskip\noindent\minipage{\linewidth}}
  {\endminipage\par\medskip}

\titleformat{\section}
{\normalfont\normalsize\bfseries}{\thesection}{0.5em}{}
\titleformat{\subsection}
{\normalfont\normalsize\bfseries\em}{\thesubsection}{0.5em}{}
\titleformat{\subsubsection}
{\normalfont\normalsize\bfseries\em}{\thesubsubsection}{0.5em}{}
\begin{document}

\begin{flushright}
\Large 

\textbf{[SSC24-XI-01]}
\end{flushright}
\begin{centering}      
\large 

\textbf{Starling Formation-Flying Optical Experiment: Initial Operations and Flight Results}\\
\vspace{0.5cm}
\normalsize 

Justin Kruger, Simone D'Amico\\
{Stanford University}\\
{496 Lomita Mall, Stanford, CA 94305}\\
{jjkruger@stanford.edu}\\ 
\vspace{0.5cm}
Soon S. Hwang\\
{NASA Ames Research Center}\\
{Moffett Field, CA 94035}\\
{soon.s.hwang@nasa.gov}

\end{centering}

\begin{centering}
    \vspace{0.5cm}
    \centerline{\textbf{ABSTRACT}}
    \vspace{0.3cm}
\end{centering}

\noindent This paper presents initial flight results for distributed optical angles-only navigation of a swarm of small spacecraft, conducted during the Starling Formation-Flying Optical Experiment (StarFOX). StarFOX is a core payload of the NASA Starling mission, which consists of four CubeSats launched in 2023. Angles-only methods apply inter-satellite bearing angles obtained by on-board cameras for navigation, increasing satellite autonomy and enabling new mission concepts. Nevertheless, prior flight demonstrations have only featured one observer and target and have relied upon a-priori target orbit knowledge for initialization, translational maneuvers to resolve target range, and external absolute orbit updates to maintain convergence. StarFOX overcomes these limitations by applying the angles-only Absolute and Relative Trajectory Measurement System (ARTMS), which integrates three novel algorithms. Image Processing detects and tracks multiple targets in images, using multi-hypothesis methods and kinematic modeling, and computes target bearing angles. Batch Orbit Determination computes initial swarm orbit estimates from bearing angle batches, via iterative batch least squares and sampling of the weakly observable target range. Sequential Orbit Determination leverages an adaptive, efficient unscented Kalman filter with nonlinear models to refine swarm state estimates over time. Multi-observer measurements shared over an intersatellite link are seamlessly fused to enable robust absolute and relative orbit determination. StarFOX flight data and telemetry presents the first demonstrations of autonomous angles-only navigation for a satellite swarm, including multi-target and multi-observer relative navigation; autonomous initialization of navigation for unknown targets; and simultaneous absolute and relative orbit determination. Relative positioning uncertainties of 1.3\% of target range (1$\sigma$) are achieved for a single observer under challenging measurement conditions, reduced to 0.6\% (1$\sigma$) with multiple observers. Results demonstrate promising performance with regards to ongoing StarFOX campaigns and the application of angles-only navigation to future distributed missions.

\begin{multicols*}{2}
\section*{INTRODUCTION}

Distributed Space Systems (DSS) can offer many advantages when compared to traditional monolithic spacecraft, including improved accuracy, coverage, flexibility, robustness, and the ability to achieve entirely new objectives \cite{damico_miniaturized_2015}.
This has led to the deployment of a variety of DSS science missions \cite{tapley_gravity_2004, krieger_tandem_2007, damico_spaceborne_2012, burch_magnetospheric_2016}, as well as the proposed application of DSS to areas such as Space Situational Awareness (SSA) \cite{holzinger_sda_2018} and In-space Servicing, Assembly and Manufacturing (ISAM) \cite{nasa_osam_2021}.
However, robust navigation for DSS remains a technological challenge.
The majority of DSS have been deployed in Earth orbit and their navigation systems therefore assume availability of external metrologies such as Global Navigation Satellite System (GNSS) signals and frequent contact with the ground.
Systems outside of Earth orbit may instead navigate via the Deep Space Network (DSN) or similar resources, but such methods impact timely decision-making for missions and are not easily scalable to future DSS.
Furthermore, navigation for non-cooperative objects such as space debris cannot be performed with GNSS.
It is therefore necessary to develop new self-contained navigation systems to enable usage of DSS in more varied scenarios, characterized by a high degree of autonomy and robustness.
Minimal technical and financial costs for associated hardware are preferred so that miniaturized technology can be leveraged.

Angles-only navigation, in which observer spacecraft obtain bearing angles to target space objects using onboard Vision-Based Sensors (VBS), is a compelling technology in this context.
VBS are already ubiquitous on modern spacecraft in the form of star trackers.
These sensors are passive and inexpensive with a high dynamic range and require minimal mass, volume, and power budgets \cite{palo_agile_2013}.
If DSS observers are also equipped with an Inter-Satellite Link (ISL), as can be implemented with typical radio frequency hardware, measurements from multiple observers can be shared and fused to improve navigation performance.
Accordingly, angles-only navigation generally requires no additional hardware even when used on small and inexpensive spacecraft.
A further benefit is that optical sensors may obtain measurements to and navigate with respect to non-cooperative targets.
However, bearing angles do not provide explicit target range information, which results in challenging observability conditions \cite{woffinden_observability_2009, sullivan_adaptive_2017}.
Target range is only weakly observable and it is often difficult to simultaneously estimate absolute and relative orbit states \cite{koenig_bod_2020}.

These aspects have motivated extensive research in angles-only navigation for spacecraft and several prior attempts to conduct angles-only navigation in flight.
These include the Orbital Express mission\cite{friend_orbital_2008}, the Advanced Rendezvous using GPS and Optical Navigation (ARGON) experiment\cite{damico_argon_2013, gaias_argon_2014}, and the Autonomous Vision Approach Navigation and Target Identification (AVANTI) experiment\cite{gaias_avanti_2017, ardaens_avanti_2018}.
Of these, ARGON and AVANTI are well-documented in literature.

ARGON (2012) was a key activity of the PRISMA mission, which was an in-orbit testbed for formation-flying and rendezvous technology developed by several European space centers.
A single observer satellite (`Tango') obtained images of a noncooperative target satellite (`Mango') in Low Earth Orbit (LEO) during an approach from 30 km to 3 km of Inter-Satellite Distance (ISD).
Navigation and control tasks were performed in a ground-in-the-loop manner using downlinked images.
Target bearing angles were provided to a batch least-squares relative orbit determination algorithm, which estimated target Relative Orbit Elements (ROE) over the measurement period.
This estimate was passed to a maneuver planner which generated guidance profiles to both achieve the rendezvous goal and improve range estimation performance.
The ground-generated plan was then uplinked.

AVANTI (2016) was conducted during the FIREBIRD mission of the German Aerospace Center.
It encompassed the rendezvous of the BIROS microsatellite and an ejected nanosatellite from 13 km to 50 m of ISD.
In comparison, AVANTI was much more autonomous and image processing, relative navigation, and maneuver planning were all performed on board.
Targets could be autonomously tracked with less reliance on orbit knowledge from two-line elements; an extended Kalman filter was used to perform ROE state estimation; and maneuver planning employed a model predictive control strategy for optimization.
Passive safety was also enforced via eccentricity/inclination-vector separation.
Notably, non-ideal conditions resulted in significant bearing angle measurement outages, as well as a degradation of attitude determination performance at closer ranges when target brightness outshone background stars in the Field Of View (FOV).

The impact of ARGON and AVANTI is felt in their successes and lessons learned.
However, they are equivalently characterized by major limitations that must be overcome to meet the needs of future missions.
Both systems did not consider multiple observers or multiple targets and focused on relative state estimation only, relying on on external knowledge of the observer's absolute orbit (e.g. from a GNSS receiver) to maintain absolute state convergence.
Additional reliances on accurate a-priori relative orbit information from the ground to initialize navigation, as well as frequent translational maneuvers to resolve the weakly observable ISD, reduced overall autonomy and efficiency.

The goal of the Starling Formation-flying Optical eXperiment (StarFOX), as described in this paper, is to remove these limitations entirely.
StarFOX is one of four experimental payloads of the NASA Starling mission managed by the NASA Ames Research Center \cite{sanchez_starling1_2018}, and is visualized in Figure \ref{fig:artist}.
Starling consists of four propulsive 6U CubeSats in LEO and aims to increase the readiness of four enabling technologies for autonomous spacecraft swarms: operations and decision-making, communications and networking, maneuver planning and execution, and absolute and relative navigation.
\begin{Figure}
\centering
\includegraphics[width=\columnwidth]{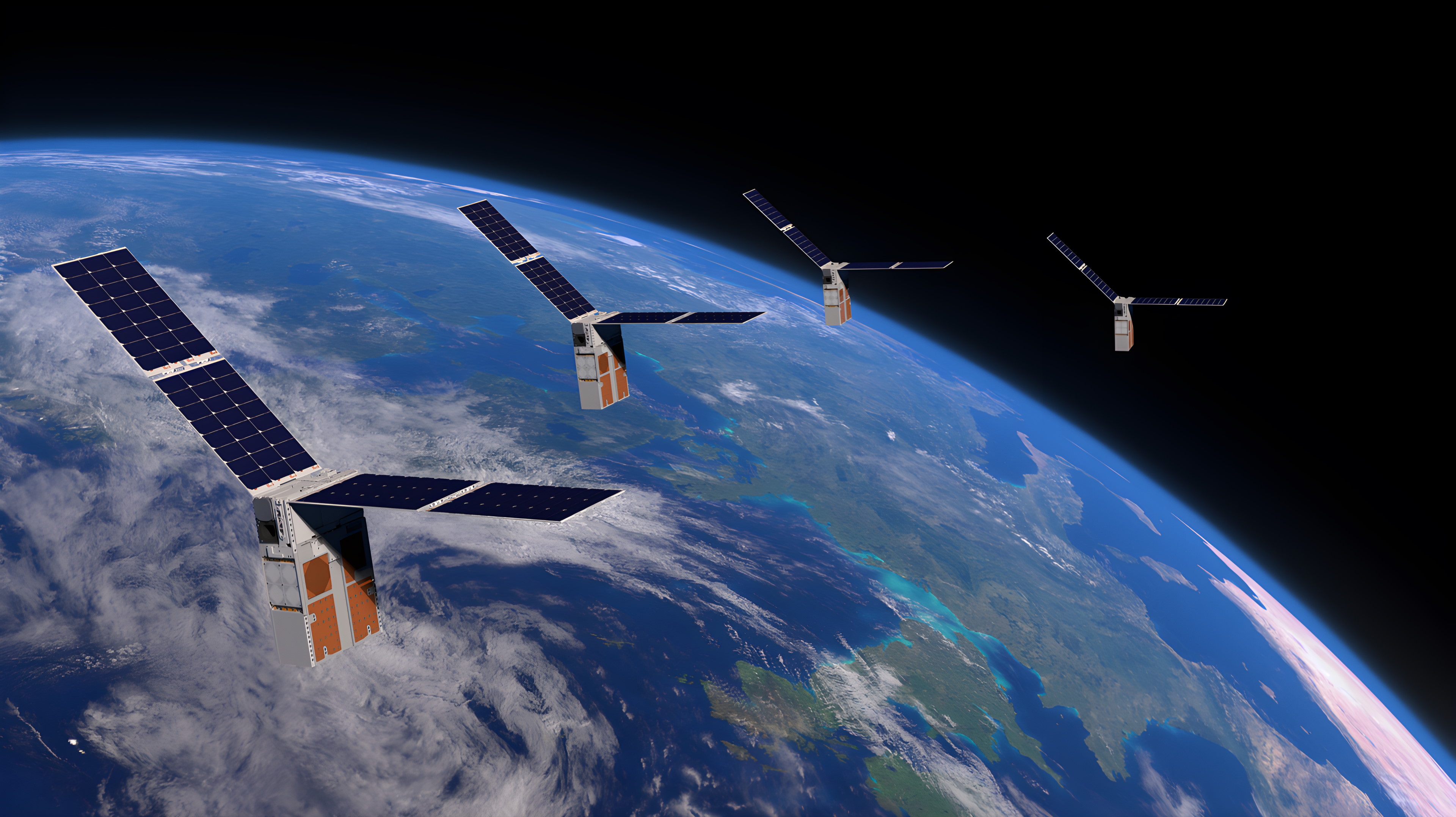}
\captionof{figure}{An artist's impression of the Starling swarm. Credit: Blue Canyon Technologies/NASA.}
\label{fig:artist}
\end{Figure}

Specifically, StarFOX applies the angles-only angles-only Absolute and Relative Trajectory Measurement System (ARTMS).
ARTMS is a self-contained architecture \cite{sullivan_generalized_2020, kruger_starfox_2023} that provides distributed, autonomous, scalable navigation capabilities for DSS orbiting an arbitrary central body, without reliance on maneuvers or external measurement sources.
Starling was launched in July 2023, and over the ensuing months, a series of StarFOX experiment blocks has explored ARTMS' flexibility for angles-only navigation in both single-observer and distributed multi-observer scenarios, using either ground-assisted or autonomous state initializations, with or without GNSS availability, in two different swarm formations.
The experiment campaign consequently builds upon ARGON and AVANTI for a more wide-ranging and ambitious demonstration of spaceborne angles-only navigation in orbit.

This paper provides an overview of StarFOX operations during the first experiment phase, as well as initial flight results obtained by 1) post-processing of flight imagery and sensor data via ARTMS flight code running on the ground, and 2) ARTMS operating on board the Starling swarm, with comparisons to pre-flight expectations.
These are the first demonstrations of autonomous angles-only navigation for a satellite swarm, including multi-target and multi-observer relative navigation; autonomous initialization of navigation for unknown targets; and simultaneous absolute and relative orbit determination.
Though initial results are promising, significant challenges were also encountered, motivating key lessons learned for future StarFOX experiments.

Following this introduction, the angles-only navigation problem is defined.
Then, ARTMS and its algorithms are described in more detail, followed by an overview of the Starling mission and hardware.
Next, the paper describes StarFOX experiment plans and operations.
Subsequently, on-ground and on-orbit results for several key navigation scenarios are presented, followed by lessons learned and conclusions. 
\section*{MATHEMATICAL BACKGROUND}
\label{sec:models}

\subsection*{Measurement Model}
ARTMS produces angles-only measurements by computing the time-tagged bearing angles to objects detected in VBS images.
Two rotating coordinate frames are defined.
First, consider the Radial/Tangential/Normal (RTN) frame of an observer spacecraft, denoted $\mathcal{R}$.
It is centered on and rotates with the observer and consists of orthogonal basis vectors $\hat{\bm{x}}^{\mathcal{R}}$ (directed along the observer's absolute position vector); $\hat{\bm{z}}^{\mathcal{R}}$ (directed along the observer's orbital angular momentum vector); and $\hat{\bm{y}}^{\mathcal{R}} = \hat{\bm{z}}^{\mathcal{R}} \times \hat{\bm{x}}^{\mathcal{R}}$ \cite{vallado_astrodynamics_2013}.
Similarly, define a frame $\mathcal{W}$ using $\hat{\bm{y}}^{\mathcal{W}}$ (directed along the observer's velocity vector); $\hat{\bm{z}}^{\mathcal{W}} = \hat{\bm{z}}^{\mathcal{R}}$; and $\hat{\bm{x}}^{\mathcal{W}} = \hat{\bm{y}}^{\mathcal{W}} \times \hat{\bm{z}}^{\mathcal{W}}$.
$\mathcal{W}$ only differs from $\mathcal{R}$ by a rotation of the observer flight path angle $\phi_f$ about $\hat{\bm{z}}^{\mathcal{R}}$ with $\phi_f \approx 0$ in near-circular orbits \cite{vallado_astrodynamics_2013}.
Finally, define the observer VBS coordinate frame $\mathcal{V}$ consisting of orthogonal basis vectors $\hat{\bm{x}}^{\mathcal{V}}, \hat{\bm{y}}^{\mathcal{V}}, \hat{\bm{z}}^{\mathcal{V}}$ where $\hat{\bm{z}}^{\mathcal{V}} = \hat{\bm{x}}^{\mathcal{V}} \times \hat{\bm{y}}^{\mathcal{V}}$ is aligned with the camera boresight.
The VBS may be pointed as necessary to keep targets in the FOV.
In the case of StarFOX, it is chosen to always point the camera boresight in the velocity or anti-velocity direction for simplicity, such that $\hat{\bm{z}}^{\mathcal{V}}$ is aligned with $\pm\hat{\bm{y}}^{\mathcal{W}}$.
This was sufficient to image targets while avoiding imaging the limb of the Earth.
Figure \ref{fig:frames} illustrates this scenario.
\begin{Figure}
\centering
\includegraphics[width=0.85\columnwidth]{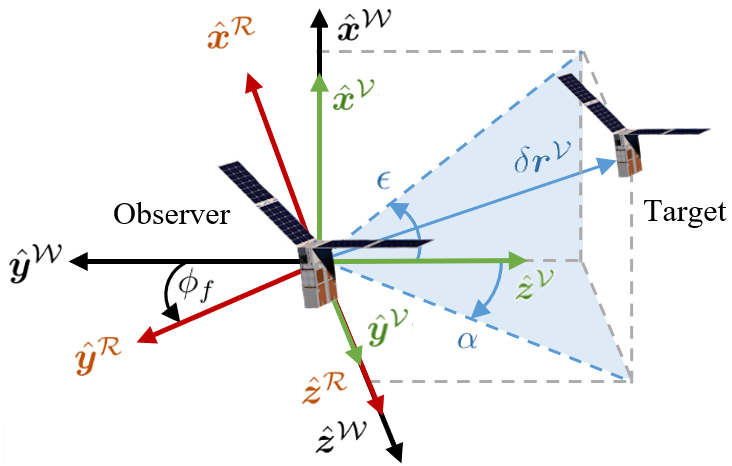}
\captionof{figure}{Definition of coordinate frames and bearing angles with VBS pointing in the $-\hat{\bm{y}}^{\mathcal{W}}$ direction.}
\label{fig:frames}
\end{Figure}

Bearing angles consist of azimuth and elevation $[\alpha, \epsilon]^\top$ and subtend the line-of-sight vector $\delta \bm{r}^{\mathcal{V}}$ from the observer to the target.
The measurement model $\bm{y}$ for the bearing angles from observer $o$ to target $t$ is \cite{sullivan_generalized_2020}
\par
\begin{equation}
\delta \bm{r}^{\mathcal{V}} = \bm{r}^\mathcal{V}_t - \bm{r}^\mathcal{V}_o = [\delta r_x^{\mathcal{V}}, \delta r_y^{\mathcal{V}}, \delta r_z^{\mathcal{V}}]^\top
\end{equation}
\begin{equation}
    \bm{y}^{\mathcal{V}}(\delta \bm{r}^{\mathcal{V}}) =
    \begin{bmatrix}
    \alpha \\
    \epsilon
    \end{bmatrix}^{\mathcal{V}}
    =
    \begin{bmatrix}
    \arcsin{(\delta r_y^{\mathcal{V}} / ||\delta \bm{r}^{\mathcal{V}}||_2)}\\
    \arctan{(\delta r_x^{\mathcal{V}} / \delta r_z^{\mathcal{V}})}
    \end{bmatrix}
\end{equation}
Measurements and states are also referenced with respect to an inertial reference frame centered on an arbitrary central body, denoted $\mathcal{I}$.
The inertial frame for StarFOX is the Earth-centered inertial J2000 frame.
Bearing angles are related to $\mathcal{I}$ by rotating $\delta \bm{r}^{\mathcal{V}}$ into $\mathcal{I}$, as per $\delta \bm{r}^{\mathcal{I}} = \ ^{\mathcal{V}} \overrightarrow{\bm{R}} ^{\mathcal{I}} \delta \bm{r}^{\mathcal{V}}$ where $^{\mathcal{V}} \overrightarrow{\bm{R}}^{\mathcal{I}}$ denotes a rotation from frame $\mathcal{V}$ into frame $\mathcal{I}$.
This rotation is generally computed by performing attitude determination using stars identified by the VBS \cite{damico_argon_2013}.
Other relevant rotations $^{\mathcal{R}} \overrightarrow{\bm{R}}^{\mathcal{I}}$ and $^{\mathcal{W}} \overrightarrow{\bm{R}}^{\mathcal{I}}$ can be computed using the observer's absolute orbit estimate, assumed to be coarsely known.
Note that in practice, VBS measurement availability is affected by optical visibility constraints such as eclipse periods.

\subsection*{System State}
Because angles-only navigation is characterized by weak observability \cite{woffinden_observability_2009}, proper selection of the state parameterization is crucial to maximize estimation accuracy and robustness.
%
%
ARTMS leverages quasi-nonsingular absolute Orbit Elements (OE) and Relative Orbit Elements (ROE).
The absolute orbit $\boldsymbol{\alpha}$ is defined as
\begin{multline}
    \bm{\alpha}
    =
    \begin{bmatrix}
    a & e_x & e_y & i & \Omega & u
    \end{bmatrix}^\top \\
    = \begin{bmatrix}
    a & e \cos \omega & e \sin \omega & i & \Omega & \omega + M
    \end{bmatrix}^\top
    \label{eq:oe}
\end{multline}
where $a, e, i, \Omega, \omega, M$ are the Keplerian orbit elements.
The relative orbit of each target detected by the onboard sensor, denoted $\delta\boldsymbol{\alpha}$, is described by the quasi-nonsingular ROE adopted by D'Amico \cite{damico_autonomous_2010}.
Each of these ROE is function of the orbit elements of the target $t$ and observer $o$ as given by
\begin{multline}
    \delta \bm{\alpha}
    =
    \begin{pmatrix}
    \delta a \\
    \delta\lambda\\
    \delta e_x\\
    \delta e_y\\
    \delta i_x\\
    \delta i_y
    \end{pmatrix}
    =
    \begin{pmatrix}
    \delta a \\
    \delta \lambda \\
    |\delta \bm{e}| \cos \phi \\
    |\delta \bm{e}| \sin \phi \\
    |\delta \bm{i}| \cos \theta \\
    |\delta \bm{i}| \sin \theta
    \end{pmatrix}\\
    =
    \begin{pmatrix}
    (a_t - a_o) / a_o\\
    (u_t - u_o) + (\Omega_t-\Omega_o) \cos i_o \\
    e_{x,t} - e_{x,o}\\
    e_{y,t} - e_{y,o}\\
    i_t - i_o\\
    (\Omega_t - \Omega_o) \sin i_o
    \end{pmatrix}
    \label{eq:roe}
\end{multline}
where $(\delta e_x, \delta e_y)$ are components of the relative eccentricity vector with phase $\phi$ and $(\delta i_x, \delta i_y)$ are components of the relative inclination vector with phase $\theta$.

These state definitions have been previously studied in literature, resulting in development of accurate analytical dynamics models \cite{koenig_new_2016, guffanti_models_2019}.
Also, these states are slowly varying and enable accurate numerical integration using Gauss's Variational Equations (GVE) with large time steps for efficient onboard orbit propagation \cite{sullivan_angles_2018}.
More importantly, the weakly observable range to each target is primarily captured by the $\delta\lambda$ term in most relative motion geometries, especially for loose satellite formations and swarms.
This allows ARTMS to maximize accuracy by applying separate state estimation techniques to different state components.
Additionally, it has been shown that the semimajor axis of the observer's orbit is strongly observable using bearing angle measurements to a single target \cite{koenig_bod_2020}.
Combined, these properties enable accurate and computationally efficient estimation algorithms with minimal reliance on a-priori information.

ARTMS also possess the capacity to estimate auxiliary state components, such as differential clock offsets and drift rates between observers, differential ballistic coefficients, and sensor biases \cite{sullivan_generalized_2020, kruger_artms_2021, keidai_artms_2022, kruger_observability_2022}.
However, the initial StarFOX experiment phase only explores orbit estimation to reduce computation costs and system complexity.
Auxiliary state estimation will be implemented as part of a StarFOX+ extended mission.

\subsection*{Dynamics Model}
\label{sec:dynamics}

ARTMS propagates the absolute orbits of system objects using fourth-order Runge-Kutta integration of the GVE.
For state $\bm{\alpha}$, the osculating OE of each spacecraft evolve according to
\begin{align}
\dot{\bm{\alpha}} &= G(\bm{\alpha})\bm{d}^{\mathcal{R}}\\
\bm{d}^{\mathcal{R}} &=
\begin{bmatrix}
    d_x^{\mathcal{R}} & d_y^{\mathcal{R}} & d_z^{\mathcal{R}}
\end{bmatrix}^\top
\end{align}
where $G \in {\mathbb{R}}^{6 \times 3}$ is the GVE matrix \cite{alfriend_spacecraft_2010} and $\bm{d}^{\mathcal{R}}$ is the perturbing acceleration expressed in $\mathcal{R}$.
Depending on the orbit regime, common perturbations include spherical harmonic gravity terms, atmospheric drag, third-body gravity and solar radiation pressure \cite{montenbruck_satellite_2012}.
In flight, ARTMS typically applies a 10x10 GM01S\cite{tapley_gravity_2004} spherical harmonic gravity model, a Harris-Priester atmosphere model with cannonball drag ($C_D \approx 0.015$), and a 30 second RK4 step.
Analytic dynamics models for the mean OE which include the effects of $J_2$ gravity are used within ARTMS when computational efficiency is paramount, e.g. during multi-hypothesis tracking and sample-based batch orbit determination.
%


With regards to relative dynamics, the ROE provide useful geometric intuition regarding target relative motion.
The curvilinear position vector of a target in the observer's RTN frame is defined as $\delta \boldsymbol{r}_{\textrm{curv}}^{\mathcal{R}} = (\delta r, a_o\Theta, a_o\Phi)$  \cite{sullivan_improved_2016}.
Here, $\delta r, \Theta, \Phi$ are target-observer differences in orbit radii, angular in-plane separations and angular out-of-plane separations respectively.
The curvilinear representation captures the effects of orbit curvature with improved accuracy and can be mapped back to rectilinear coordinates via
\begin{equation}
\delta \bm{r}_{\textrm{rect}}^{\mathcal{R}}
=
\begin{bmatrix}
(a_o + \delta r) \cos{\Theta} \cos{\Phi} - a_o \\
(a_o + \delta r) \sin{\Theta} \cos{\Phi} \\
(a_o + \delta r) \sin{\Phi}
\end{bmatrix}
\label{eq:mapping}
\end{equation}
As first shown by D'Amico\cite{damico_autonomous_2010}, there exists a linear mapping between the ROE and $\delta \boldsymbol{r}_{\textrm{curv}}^{\mathcal{R}}$, defined as
\begin{equation}
\delta \boldsymbol{r}^{\mathcal{R}}
\approx
a_o
\begin{bmatrix}
\delta a - \delta e_x \cos u_o\\
-1.5 \delta a + \delta \lambda + 2 \delta e_x \sin u_o - 2 \delta e_y \cos u_o\\
\delta i_x \sin u_o - \delta i_y \cos u_o
\end{bmatrix}
\label{eqn:EROEmap}
\end{equation}
The mapping was later extended to eccentric orbits by defining eccentric ROE\cite{sullivan_nonlinear_2017} $\delta \bm{\alpha}^* = [\delta a, \delta \lambda^*, \delta e_x^*, \delta e_y^*, \delta i_x, \delta i_y]^\top$, which revert to the quasi-nonsingular ROE for $e_o \approx 0$.

Figure \ref{fig:ellipse} presents relative motion in RTN for this mapping for small ISD.
Components of oscillatory motion produced by the target's separation and relative orbit are shown in black, possessing the same frequency as the orbit.
Components of oscillatory motion produced by orbit eccentricity are shown in red, acting at twice the frequency of the orbit.
$\delta a$ and $\delta \lambda^*$ capture mean offsets in the radial and along-track directions respectively; magnitudes of $\delta e^*$ and $\delta i$ correspond to magnitudes of oscillations in the RT and RN planes respectively; and phases of $\delta e^*$ and $\delta i$ dictate the orientation and aspect ratio of the tilted ellipse in the RN plane. The eccentricity of the observer's orbit superimposes additional offsets and higher-frequency oscillations in the RT and RN planes.
\begin{Figure}
\centering
\includegraphics[width=\columnwidth]{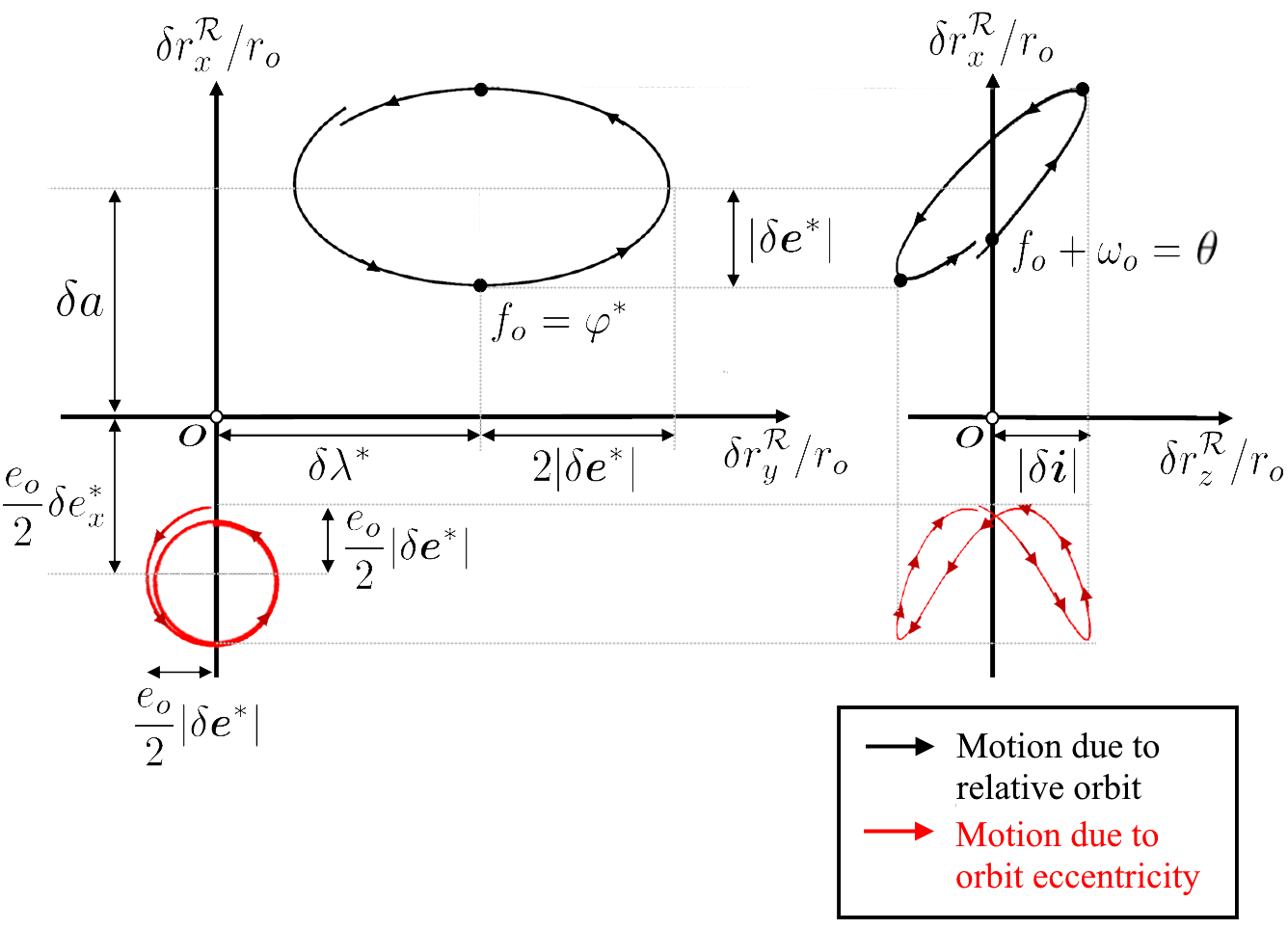}
\captionof{figure}{Components of target relative motion in the $\hat{\bm{x}}^{\mathcal{R}}$-$\hat{\bm{y}}^{\mathcal{R}}$ (RT) and $\hat{\bm{x}}^{\mathcal{R}}$-$\hat{\bm{z}}^{\mathcal{R}}$ (RN) planes \cite{sullivan_generalized_2020}.}
\label{fig:ellipse}
\end{Figure}
\section*{NAVIGATION ARCHITECTURE}
\label{sec:architecture}

To simplify subsequent discussions, the following terminologies are adopted.
The ``observer'' refers to the spacecraft hosting the instance of ARTMS being discussed.
A ``remote observer'' is another spacecraft hosting ARTMS providing measurements over the ISL.
The ``local subsystem'' includes an observer and all its ``targets'', which are the Resident Space Objects (RSO) detected by the onboard VBS.
The ``system'' (or DSS) refers to the entire distributed space system, consisting of all involved observers and targets.
Targets may be non-cooperative objects or spacecraft which do not actively assist navigation, or may be cooperative remote observers themselves.
Figure \ref{fig:swarm} presents a notional illustration of a four-spacecraft DSS using ARTMS.
\begin{Figure}
\centering
\includegraphics[width=\columnwidth]{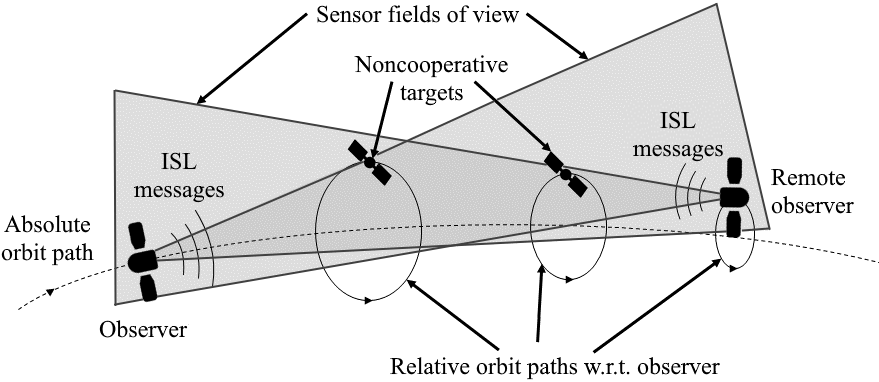}
\captionof{figure}{Notional illustration of ARTMS observers and targets for a four-spacecraft system (not to scale).}
\label{fig:swarm}
\end{Figure}

\begin{figure*}[hbt]
\centering
\includegraphics[width=0.85\textwidth]{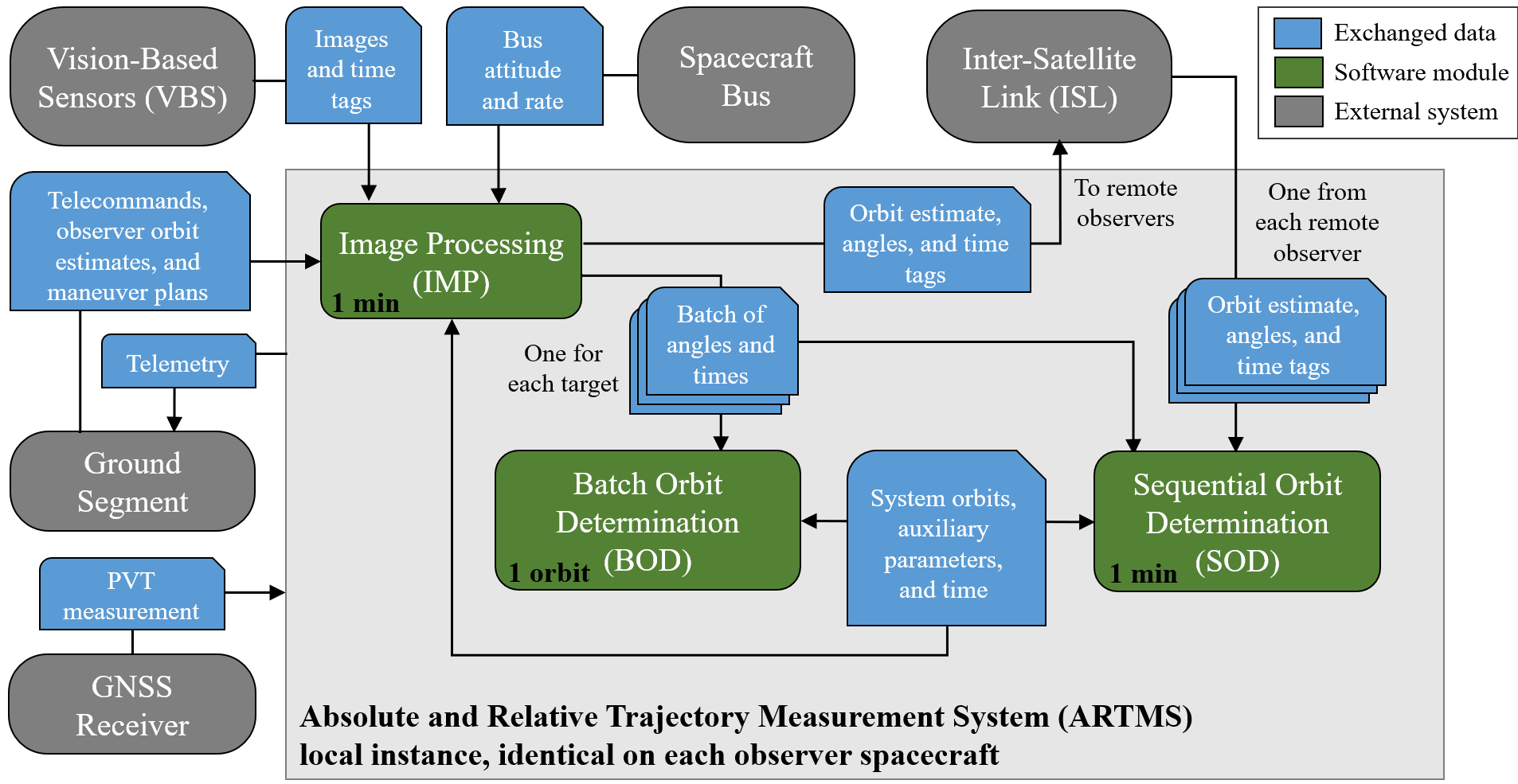}
\caption{A high-level overview of the ARTMS flight architecture.}
\label{fig:architecture}
\end{figure*}

A high-level overview of ARTMS as implemented in StarFOX flight software is shown in Figure \ref{fig:architecture}.
The architecture of three core modules based on algorithms recently developed at Stanford's Space Rendezvous Laboratory (SLAB): IMage Processing (IMP) \cite{kruger_imp_2021}, Batch Orbit Determination (BOD) \cite{koenig_bod_2020}, and Sequential Orbit Determination (SOD) \cite{sullivan_generalized_2020}.
The VBS provides time-tagged raw images, which are processed to obtain inertial bearing angles to target RSO.
The ISL communicates orbit estimates and angles measurements between observers in the DSS, which allows ARTMS to perform distributed multi-observer navigation.
The ground segment provides telecommands, maneuver plans and orbit estimates to each observer in the DSS, and receives ARTMS telemetry.
If the spacecraft is equipped with a GNSS receiver, it optionally provides Position/Velocity/Time (PVT) navigation solutions to replace less timely orbit estimates from the ground.
This constitutes the only a-priori information needed, i.e. to initialize, each observer must possess a coarse estimate of its own orbit at a single epoch.

The IMP module uses a coarse estimate of the observer's orbit and VBS images to produce batches of bearing angles and corresponding uncertainties to all visible targets without any a-priori relative orbit knowledge.
Relative orbit information from SOD can be leveraged when available to reduce computation cost.
IMP measurement batches are provided to the BOD and SOD modules.
Additionally, IMP sends the orbit estimate and bearing angles to the ISL for transmission.

The BOD module uses the coarse estimate of the observer's absolute orbit and the batches of bearing angles provided by IMP to compute orbit estimates for all spacecraft in the local system (including itself and all targets observed by onboard cameras).
This DSS state estimate is provided to SOD for initialization and fault detection.

The SOD module uses the state estimate from BOD to initialize a navigation filter that continuously estimates the orbits of all spacecraft in the local system as well as auxiliary parameters (e.g. ballistic coefficients or differential clock offsets).
SOD seamlessly fuses measurements from IMP and from remote observers communicated over the ISL.
SOD state estimates are provided to the ground and also to IMP to reduce the computation cost of tracking detected targets.

Overall, ARTMS provides real-time orbit estimates for the host spacecraft and each target detected by the onboard sensor. 
The only hardware requirements posed on the spacecraft are that it must have a VBS and an ISL.
ARTMS navigation is primarily autonomous in that minimal a-priori information is required and no external measurement sources are needed.
The ISL enables cooperative navigation, and the system is distributed in the sense that each observer only navigates for its local subsystem, which may be a subset of the complete DSS.

\subsection*{Image Processing}

The objective of the IMP module is to produce batches of time-tagged bearing angle measurements to each target, using images from the onboard VBS.
This is accomplished in two phases.
First, each incoming image is processed and reduced to a set of inertial bearing angles that may correspond to resident space objects.
Second, these candidate bearing angles are used to track known targets and detect new targets using an approach inspired by Multi-Hypothesis Tracking (MHT) \cite{vo_mtt_1999}.
Figure \ref{fig:exampleimage} presents an example image input, with typical sample rates of 60-120 seconds during StarFOX in LEO, equating to 50-100 images per orbit.
\begin{Figure}
\centering
\includegraphics[width=\columnwidth]{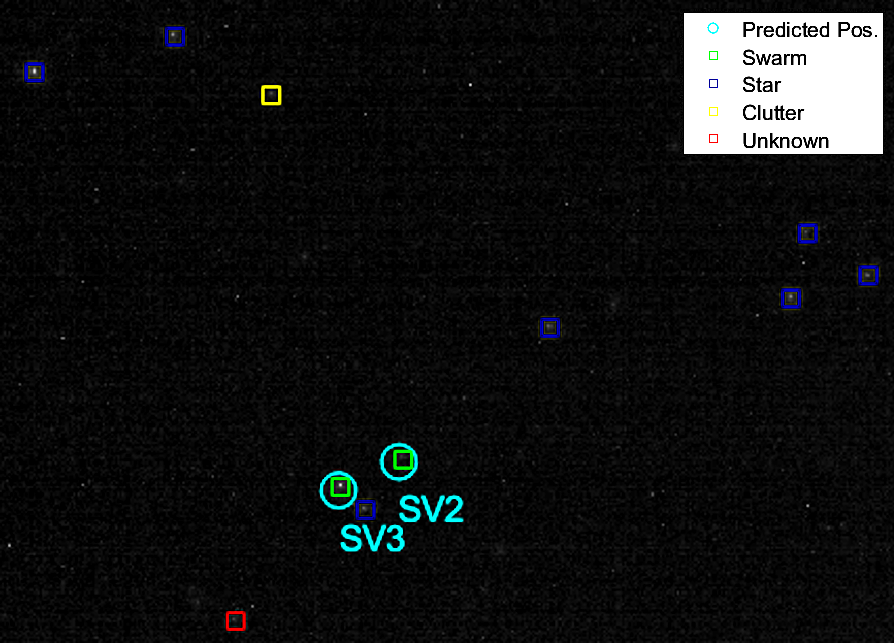}
\captionof{figure}{A Starling flight image from 12/16/2023 with point sources detected and identified by IMP.}
\label{fig:exampleimage}
\end{Figure}

The first phase of IMP uses a set of state-of-the-art algorithms.
An iterative weighted center of gravity or fast Gaussian fitting centroiding algorithm \cite{wan_centroiding_2018} is used to simplify the raw image into a list of pixel cluster centroids.
Centroids are converted to unit vectors in the VBS frame using the calibrated sensor model.
Next, the Pyramid star identification algorithm \cite{mortari_pyramid_2004} is applied to remove Stellar Objects (SO) from the list of pointing vectors.
Uncatalogued SO are detected by considering objects with unchanging inertial unit vectors between images.
Similarly, camera hotspots are removed by considering objects with unchanging pixel coordinates.
The VBS attitude is computed from the pointing vectors to identified stars in the inertial and sensor frames using the q-method \cite{wertz_spacecraft_2012}.
Attitude is computed using the same sensor which provides the bearing angle measurements to minimize calibration errors.
The remaining minimalistic set of inertial unit vectors (and equivalent bearing angles) likely corresponds to known targets or other unknown objects in the FOV.

In the second phase, these measurement candidates must be consistently assigned to new targets or tracked targets without requiring a-priori relative orbit knowledge.
To accomplish this, IMP employs the new Spacecraft Angles-only MUltitarget tracking System (SAMUS) \cite{kruger_imp_2021}, which has only two key requirements: 1) a coarse estimate of the observer's absolute orbit is available, and 2) targets must not perform unknown translational maneuvers during the tracking period.
SAMUS is valid for eccentric orbits and has been specifically designed to meet the constraints of risk-averse angles-only navigation in space, i.e. to achieve close to 100\% measurement assignment precision with low measurement frequencies and limited computational resources.

SAMUS applies the core concept of MHT in that as measurements arrive, many simultaneous hypotheses are maintained as to how they can be associated into target tracks.
The algorithm converges towards the correct hypothesis over time, to enhance robustness of its outputs.
MHT is chosen as a basis because it is mature and demonstrably accurate, with its primary disadvantage being the need to frequently and heuristically prune hypotheses for real-time computation \cite{blackman_mht_2004}.
To overcome this, SAMUS applies domain-specific knowledge to develop precise pruning criteria.

Recall Equation \ref{eq:mapping}, which maps the ROE to a target's curvilinear position vector in the observer's RTN frame.
In the map, $u$ is the only quickly-varying term, while all other terms vary slowly in the presence of perturbations. 
Thus, target motion is described by periodic, parametric functions with known form.
Even if the specific ROE in Equation \ref{eq:mapping} are unknown, its form provides expectations regarding target motion that can be leveraged.

%
%

Given a set of past bearing angles measurements in a track and corresponding estimates of the observer's absolute orbit, Equation \ref{eq:mapping} can be rearranged into a pair of separable linear systems in azimuth and elevation \cite{kruger_imp_2021}, as in
\begin{align}
\begin{bmatrix}
\epsilon\\
\alpha
\end{bmatrix}^\mathcal{R}
& \approx
\frac{r_o}{a_o}
\begin{bmatrix}
x_1 - x_2 \cos (u_o - x_3)\\
x_4 + x_5 \sin(u_o - x_6)
\end{bmatrix}
\label{eqn:fit}
\end{align}
where $x_{1,...,6}$ are scaled ROE equivalents in bearing angle space \cite{kruger_imp_2021}.
Terms $r_o, a_o, f_o, e_o, \omega_o$ are computed from the observer orbit estimate.
If a track consists of at least three bearing angles at different times, the six unknowns $x_{1,...,6}$ can be solved for via least squares.
Measurements in future epochs can then be predicted by propagating the observer orbit estimate and applying the fitted model, and goodness of fit can be assessed via fitting residuals.

To assess which hypotheses are physically reasonable, SAMUS applies a set of kinematic rules\cite{kruger_imp_2021} derived from the parametric motion model.
Only tracks which pass all rules are propagated.
Briefly, the rules are summarised as: 1) track velocities must be below a set maximum, 2) track velocities must be consistent over time, 3) tracks should generally not feature acute angles, 4) tracks should turn in a consistent direction, 5) new measurements must be close to the predicted measurement.
Their application greatly increases efficiency of MHT by preventing formation of unlikely tracks.

When multiple tracks pass all rules, SAMUS scores propagated tracks via ten criteria which assess how well each fulfills the expectations of Equation \ref{eq:mapping}, the predicted measurement, and their prior motion.
%
%
In contrast to traditional MHT methods $-$ which often rely on a single Mahalanobis distance metric for scoring $-$ SAMUS aims to be more robust.
Often, target tracks intersect or are in close proximity in the image plane, or motion between images is on the order of VBS noise.
A single scoring metric is therefore not robust.
By using a larger set of metrics, consensus supports the correct choice over time, even if some temporarily support incorrect hypotheses.
Additionally, scoring does not require probabilistic estimates of false alarm densities or target decay rates, which are not easily obtainable for spacecraft.

To initialize new tracks, SAMUS employs the Density-Based Spatial Clustering of Applications with Noise (DBSCAN) algorithm \cite{dbscan_1996}.
DBSCAN clusters require $\geq n_D$ points within small radius $\epsilon_D$.
Because targets are in similar orbits to observers, their velocities compared to other objects in the FOV are low.
Previously untracked targets are initialized by applying DBSCAN to the merged set of unidentified measurements from the past several images, and applying the SAMUS kinematic rules to found clusters.

Finally, tracking is often interrupted by orbit eclipse periods.
To connect shorter tracks on either side of an eclipse, the aforementioned linear system fit is computed for every possible set of paired tracks.
The combination of compatible pairs which produces the least fitting residuals is chosen as output.

SAMUS is also able to cooperate with SOD and apply target state knowledge (if available) for improved computation cost.
Target state estimates from SOD are propagated into the current epoch to provide predicted track measurements.
The kinematic rules are replaced by a validity region around the prediction, computed via an unscented transform of the target state covariance.
The Mahalanobis distance between predicted and assigned measurement is employed for track scoring.
Figure \ref{fig:samus} presents an overview of core SAMUS operations.

\begin{Figure}
\centering
\includegraphics[width=\textwidth]{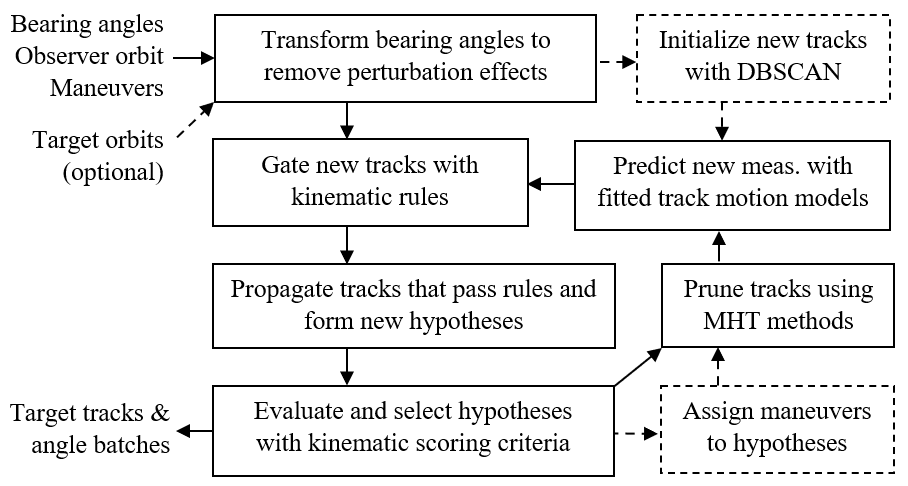}
\captionof{figure}{SAMUS algorithm summary and core sequence of operations.}
\label{fig:samus}
\end{Figure}

\subsection*{Batch Orbit Determination}

The BOD module must be able to produce orbit estimates for the local swarm with sufficient accuracy to initialize the SOD module using only a coarse estimate of the observer's orbit and batches of bearing angles to each target from IMP.
Nominally, the module operates on measurement batches of [50, 200] measurements across 1-2 orbits, per target.
State estimation is accomplished using an algorithm that separately estimates the relative orbits of each target while simultaneously refining the observer's semimajor axis estimate\cite{koenig_bod_2020}.
When used in LEO, the algorithm applies a fully analytical dynamics model including the earth oblateness $J_2$ perturbation developed by the authors to minimize computation cost \cite{koenig_new_2016}.

%
For each target, the system state is divided into estimated components $\bm{x}_{\textrm{est}}$ and components known a-priori $\bm{x}_{\textrm{prior}}$.
For StarFOX, $\bm{x}_{\textrm{est}} = [a, \delta a, \delta e_x, \delta e_y, \delta i_x, \delta i_y]$ because these state components are more strongly observable, with $\bm{x}_{\textrm{prior}} = [e_x, e_y, i, \Omega, u]$.
To robustly determine the weakly observable $\delta \lambda$, a one-dimensional family of state estimates is computed for user-specified values of $\delta\lambda$; the set of possible $\delta \lambda$ values is selected based on system limitations such as sensitivity of the VBS.

First, to initialize $\bm{x}_{\textrm{est}}$ for each $\delta \lambda$ sample, $a$ is extracted from the observer's orbit estimate and all other ROE are set to zero.
Iterative batch least squares refinement is then performed.
Define the batch of true angle measurements from IMP as $\bm{z}_{\textrm{meas}}$ for measurement epochs $\bm{t}$, and a batch of modeled angle measurements $\bm{z}_{\textrm{model}}$, computed by propagating the estimated target state to each epoch in $\bm{t}$.
The difference between measured and modeled angles is
\begin{equation}
    \Delta \bm{z} = \bm{z}_{\textrm{meas}} - \bm{z}_{\textrm{model}}(\bm{x}(t_{\textrm{est}}), \bm{t})
\end{equation}
where $t_{\textrm{est}}$ is the BOD estimation epoch. 
Subsequently, the state estimate update
\begin{align}
    \Delta \bm{z} &= \bm{S}_{\textrm{est}}(\bm{x}(t_{\textrm{est}}), \bm{t}) \Delta \bm{x}_{\textrm{est}}\\
    \bm{x}_{\textrm{est}} &\leftarrow \bm{x}_{\textrm{est}} + \Delta \bm{x}_{\textrm{est}}
\end{align}
can be computed, where $\bm{S}_{\textrm{est}}(\bm{x}(t_{\textrm{est}}), \bm{t})$ is a sensitivity matrix containing the partial derivatives of measurements with respect to state components, computed via linearization of the bearing angle measurement model.
The unknown $\Delta \bm{x}_{\textrm{est}}$ is solved for via least squares.
Refinement continues until an iteration limit is reached or $\Delta \bm{x}_{\textrm{est}}$ is smaller than a convergence threshold.

The output state estimate is selected as the $\delta \lambda$ candidate which produced the smallest measurement residual vector $\Delta \bm{z}$.
Figure \ref{fig:BOD} shows the norms of the converged measurement residual vectors for each candidate value of $\delta\lambda$ in a single test case of the BOD algorithm.
%
%
Next, the measurement noise matrix for each measurement (denoted $\bm{R}_{vbs}$) is estimated using the measurement residuals corresponding to the final state estimate.
This estimation of sensor noise using the post-fit measurement residuals allows operation even when an accurate a-priori model of sensor noise is unavailable.

Finally, the covariance $\bm{P}_{\textrm{est}}$ for estimated state components $\bm{x}_{\textrm{est}}$ is computed as given by
\begin{equation}
    \bm{P}_{\textrm{est}} = \bm{Y}^*_{\textrm{est}} (N\bm{R}_{\textrm{vbs}} + \bm{Y}_{\textrm{prior}} \bm{P}_{\textrm{prior}} \bm{Y}^\top_{\textrm{prior}}) \bm{Y}^{*\top}_{\textrm{est}} 
\end{equation}
where $\bm{Y}^{*}_{\textrm{est}}$ is the pseudoinverse of the measurement sensitivity matrix for estimated state components, $\bm{Y}_{\textrm{prior}}$ is the measurement sensitivity matrix for a-priori information, $\bm{P}_{\textrm{prior}}$ is the uncertainty of the a-priori information, and $N$ is the number of provided bearing angle measurements.
This formulation allows the BOD module to seamlessly transition between domains where uncertainty is driven by sensor performance and by errors in the a-priori information.
The ROE in $\bm{x}_{\textrm{est}}$ for each target are then appended to $\bm{x}_{\textrm{prior}}$ and the mean refined observer semimajor axis, forming a complete orbit estimate for the local swarm.

\begin{Figure}
\centering
\includegraphics[width=\columnwidth]{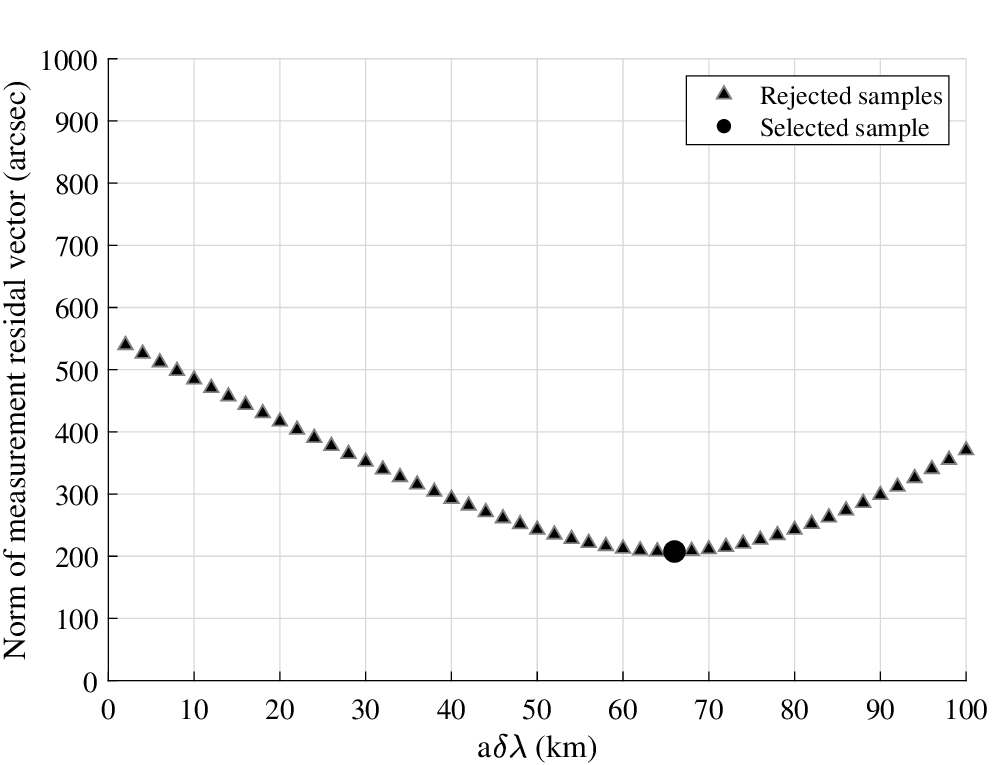}
\captionof{figure}{Behavior of converged measurement residuals for different $\delta \lambda$ samples in the BOD module.}
\label{fig:BOD}
\end{Figure}

\subsection*{Sequential Orbit Determination}

The SOD module continually refines estimates of the orbits of all spacecraft in the local swarm as well as auxiliary parameters (e.g. differential clock offsets) by seamlessly fusing measurements from all observers transmitted over the ISL.
SOD is based on a new adaptive, efficient Unscented Kalman Filter (UKF) with fully nonlinear dynamics and measurement models.
In comparison to an extended Kalman filter, the UKF preserves higher order moments in the probability distribution, which enables maneuver-free convergence using angles-only measurements from a single observer, without excessive computational cost \cite{sullivan_generalized_2020}.

Three additional features are included to maximize performance using measurements from multiple observers.
First, adaptive process noise estimation is used to improve convergence speed and robustness to errors in the dynamics model \cite{sullivan_adaptive_2017}.
Second, the state definition is organized in a way that exploits the structure of the Cholesky factorization to reduce the number of calls to the orbit propagator by almost a factor of two \cite{stacey_autonomous_2018}.
Third, measurements from remote observers are assigned to tracked targets using selection criteria based on the Mahalanobis distances between the estimated bearing angles to each target and each candidate measurement.
Let $\sigma_{jk}$ denote the Mahalanobis distance between the $j$th measurement from the remote observer and the predicted measurement to the $k$th target tracked by the local observer (which accounts for all relevant state uncertainties).
To minimize erroneous assignments, the $j$th measurement from the remote observer is assigned to the $k$th target if three conditions are satisfied:\\
\indent 1) $\sigma_{jk} \le \epsilon_{assign}$\\
\indent 2) $\sigma_{lk} \ge \epsilon_{ambig}$  $\forall \; l \ne j$\\
\indent 3) $\sigma_{jp} \ge \epsilon_{ambig}$  $\forall \; p \ne k$\\
where $\epsilon_{assign}$ and $\epsilon_{ambig}$ are user-specified parameters that satisfy $\epsilon_{ambig} > \epsilon_{assign} > 0$.
These conditions ensure 1) the measurement is close to the modeled measurement of the target using the current state estimate, 2) there is no other candidate measurement that fits the estimated state of the target, and 3) there is no other target with a state estimate that fits the measurement.

Figure \ref{fig:assignment} includes conceptual illustrations of four possible cases of modeled and observed measurements from a remote observer which (from left to right) show all conditions satisfied and violations of Condition 1, Condition 2, and Condition 3, respectively.
Together, these conditions ensure that measurements are only assigned when the observed and modeled measurements uniquely agree with a statistical certainty determined by the values of $\epsilon_{assign}$ and $\epsilon_{ambig}$.
The values of these parameters should be selected based on the expected number of targets, relative motion geometry, sensor noise, and available orbit knowledge for general swarming missions.
However, for scenarios similar to StarFOX, the authors have found that setting $\epsilon_{assign} = 3$ and $\epsilon_{ambig} = 6$ provide robust measurement assignment performance.
\begin{Figure}
\centering
\includegraphics[width=\columnwidth]{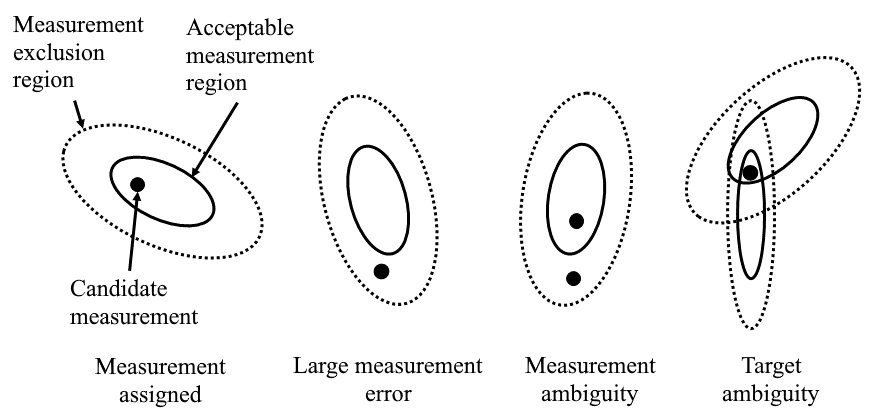}
\captionof{figure}{Conditions in which all measurement assignment criteria are satisfied (leftmost) and conditions that violate each criteria (right).}
\label{fig:assignment}
\end{Figure}

Finally, to aid robustness and error recovery, the SOD module performs autonomous health checks for the current state estimate after each filter update and attempts to automatically re-initialize if necessary.
Re-initialization is performed if: 1) the Mahalanobis distance between the most recent BOD relative state estimate and current SOD relative state estimate is above a threshold, 2) the number of successive filter update steps which did not feature a measurement update is above a threshold, 3) the number of successive filter measurement update steps which featured a large pre-fit measurement residual is above a threshold, 4) the SOD state estimate is outside a valid range (as specified by the user).
To re-initialize, the SOD module may either use ground-provided orbit estimates or the most recent BOD state estimate.
\section*{ON-ORBIT TESTBED}
\label{sec:testbed}

StarFOX is a core payload of the NASA Starling swarm mission, which consists of four 6U CubeSats.
Starling was initially proposed in 2018 \cite{sanchez_starling1_2018} as a testbed for autonomous swarming technologies, with a focus on four separate experiment payloads: autonomous networking, autonomous decision-making, autonomous maneuvering, and autonomous navigation.
Development of the StarFOX navigation experiment began in 2019, and integration and testing of the ARTMS flight software continued until the end of 2022\cite{kruger_starfox_2023}.
The swarm was launched into an approximately sun-synchronous low Earth orbit on July 17, 2023.
The primary mission phase was expected to allow for approximately six months of experiments, with the spacecraft transitioning through multiple formations.
Nominal experiment operations commenced in November 2023 and continued until May 2024.

\subsection*{Bus Hardware}

For StarFOX, each CubeSat carries two Blue Canyon Technologies Nano Star Trackers\cite{palo_agile_2013} (NST) aligned in antiparallel directions, either of which may collect imagery.
Parameters of the NST are provided in Table \ref{tab:cam}.
During each experiment, one NST is chosen as the experiment camera and is pointed in the velocity or anti-velocity direction, to image swarm members.
Images were provided to ARTMS with a cadence of 60-120 seconds.
The other NST provides a backup spacecraft attitude solution.
\begin{center}
\fontsize{8}{7.2}\selectfont
{\tabulinesep=1.1mm
\captionof{table}{Intrinsic parameters of the NST.}
\label{tab:cam}
\begin{tabu}{|p{3.0cm}|p{2.0cm}|}
    \hline
    Parameter & Value \\ \hline
    Image Size (pixels) 		& 1280 $\times$ 1024 	\\
    FOV	 ($\degree$)		& 12 $\times$ 10		\\
    Pixel Size ($\mu$m)	& 5.3			\\
    Focal Length (mm)	& 30			\\
    Pixel Space & 12-bit (grayscale)\\
    \hline
\end{tabu}}
\end{center}

StarFOX also receives inputs from the onboard GPS receiver, which optionally provides PVT solutions, and from the ISL, through CesiumAstro S-band software-defined radios.
If the ISL is active, each spacecraft transmits a message to all other swarm members every 60 seconds and listens for messages from all other members.

The payload processor for StarFOX is a Xiphos Q7S running at approximately 700 MHz.
This payload processor is separate from the primary flight processor and similarly, the ARTMS flight software is kept separate from the higher-level core flight software, to avoid impacting nominal mission operations.
Finally, each satellite also possesses a cold gas propulsion system with four separate nozzles, used to conduct swarm phasing and reconfiguration maneuvers.

Figure \ref{fig:bus} shows the four Starling spacecraft during integration at the NASA Ames Research Center.
The spacecraft are denoted SV1, SV2, SV3 and SV4 or more affectionately as Inky, Blinky, Pinky, and Clyde.
\begin{Figure}
\centering
\includegraphics[width=\columnwidth]{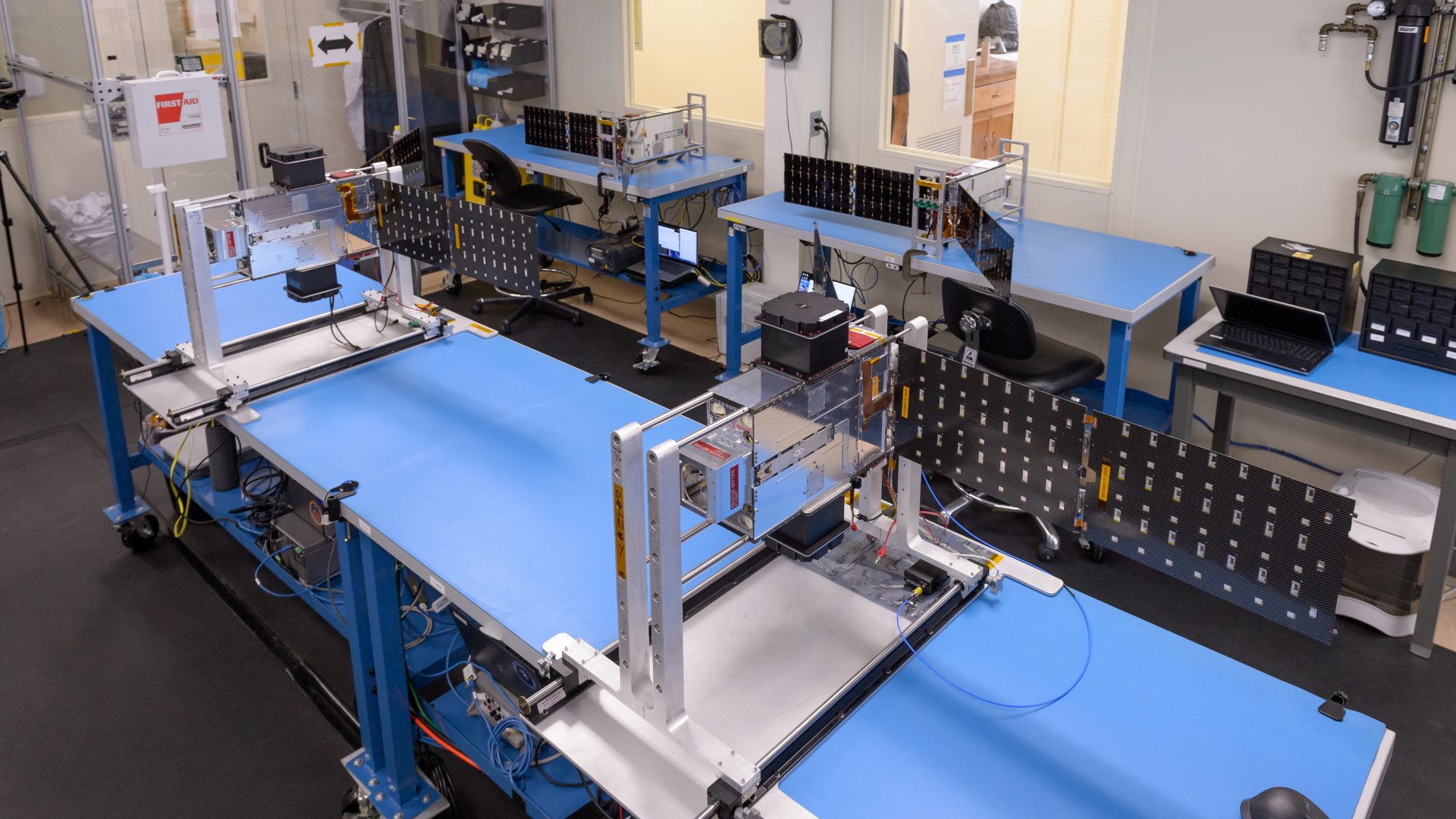}
\captionof{figure}{The four Starling CubeSats during integration. Credit: NASA/Dominic Hart (2022).}
\label{fig:bus}
\end{Figure}

\subsection*{Swarm Geometry}

Swarm geometry for the Starling mission was determined by the differing needs of flight experiments and in-flight safety considerations. For StarFOX, the following requirements were specified:
\begin{enumerate}[noitemsep]
    \item The ISD between swarm members shall not exceed crosslink communication range.
    \item The ISD shall not exceed the limit for detection of targets in star tracker images.
    \item ISD shall be greater than a minimum safe distance.
    \item Bearing angle measurements of targets shall not remain constant (i.e., observable relative motion).
    \item Relative motion between spacecraft shall include unique variations in the radial/normal directions.
    \item Relative motion ellipses shall lie entirely within the VBS FOV when the VBS boresight is aligned with the (anti)-velocity direction.
\end{enumerate}
In order to provide varying geometry for StarFOX, two distinct formations were used: an in-train (IT) formation and a passive safety ellipse (PSE) formation.
The IT formation features relatively little relative motion, with the swarm primarily separated in the velocity direction.
The PSE formation provides additional relative motion (to aid angles-only observability) as well as passive safety via relative eccentricity/inclination vector separation\cite{montenbruck_e/i-vector_2006}.
The swarm was initially placed into the IT formation, then transitioned into the PSE formation in March 2024.

Table \ref{tab:orbits} presents OE for SV4, and ROE for SV2 and SV3 (with respect to SV4), following the conventions of Equations \ref{eq:oe} and \ref{eq:roe}.
SV1 was not involved in primary StarFOX operations.
Figure \ref{fig:formations} displays the relative motion geometry of the swarm in the radial-tangential and radial-normal planes of SV4, in curvilinear coordinates.
%

\begin{center}
\fontsize{8}{7.2}\selectfont
{\tabulinesep=1.1mm
\captionof{table}{Orbit elements of the swarm.}
\label{tab:orbits}
\begin{tabu}{|p{0.5cm}|p{6.7cm}|}
    \hline
    ID & In-Train Formation (on 02/05/24 at 00:00:00 UTC)\\
    \hline
    SV4 &  $\bm{\alpha} = [6945\; \textrm{km}, 0.0007, 0.0014, 99.4\degree, -152.3\degree, -41.8\degree]$\\
    SV2 & $\delta \bm{\alpha} = [21, -124350, 110, 202, 79, 1005]$ m\\
    SV3 & $\delta \bm{\alpha} = [-1, -79328, 42, 452, 36, 827]$ m\\
    \hline
    ID & Passive Safety Ellipse (on 05/08/24 at 00:00:00 UTC)\\
    \hline
    SV4 &  $\bm{\alpha} = [6943\; \textrm{km}, 0.0001, 0.0016, 99.4\degree, -38.0\degree, -153.0\degree]$\\
    SV2 & $\delta \bm{\alpha} = [23, -52266, -471, -523, 35, 1811]$ m\\
    SV3 & $\delta \bm{\alpha} = [-30, 155320, -297, -315, 16, 1144]$ m\\
    \hline
\end{tabu}}
\end{center}

\begin{Figure}
\centering
\includegraphics[width=\columnwidth]{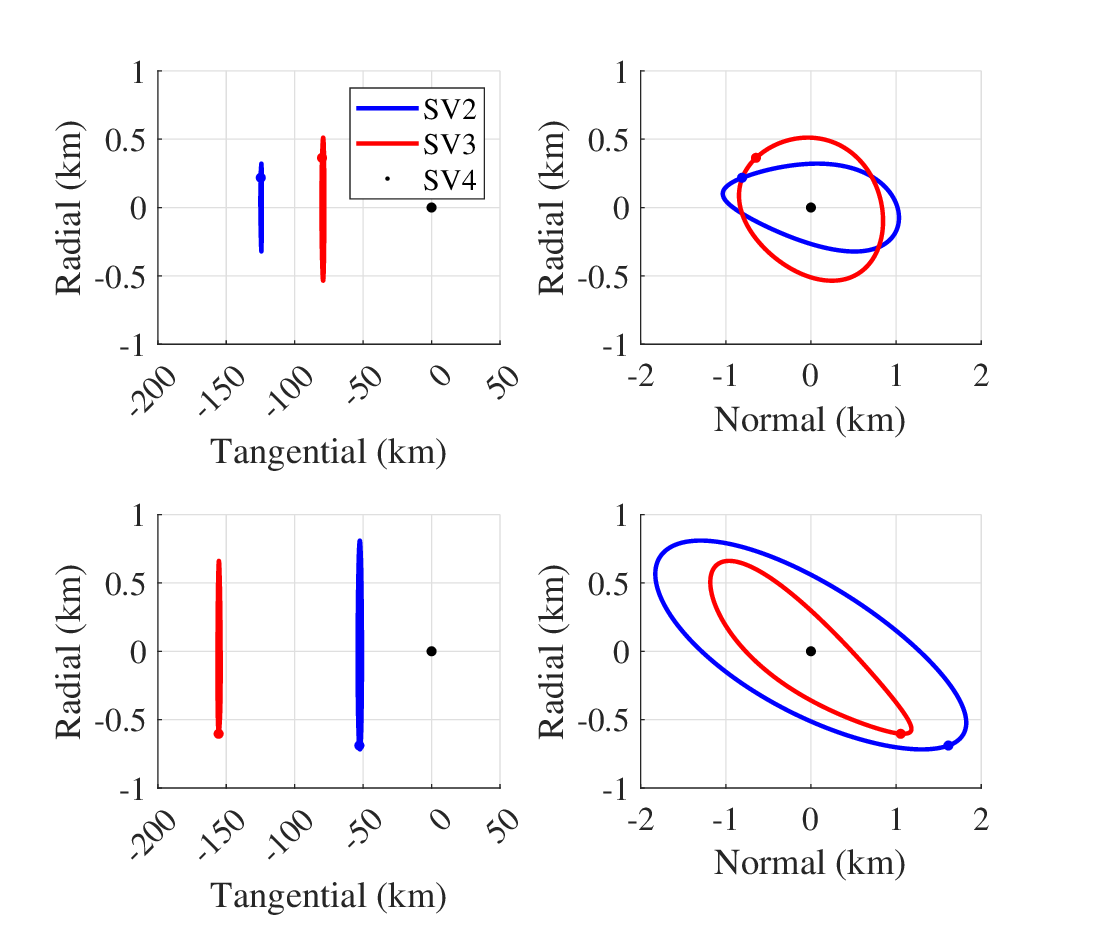}
\captionof{figure}{Swarm relative motion in the IT formation (upper) and PSE formation (lower).}
\label{fig:formations}
\end{Figure}
\section*{FLIGHT OPERATIONS}
\label{sec:operations}

The initial Starling mission was divided into six phases: 1) launch and deployment, 2) commissioning, 3) formation acquisition, 4) in-train formation, 5) passive safety ellipse formation, 6) extension.
This paper concerns StarFOX experiment operations during Phases 4) and 5).
Ongoing communication and propulsion difficulties affecting SV1 meant that it could not be reliably included in StarFOX experiments, and furthermore, its relative orbit was not consistently visible from the other swarm members.
Experiments were therefore planned using SV2, SV3 and SV4 only.

\subsection*{Experiment Planning}

In its current form, StarFOX proposes four primary navigation modes, as enabled by ARTMS:
\begin{enumerate}[noitemsep]
\item Single observer, partially autonomous: initial target relative orbit estimates are provided by the ground and refined on board by IMP + SOD. GPS is used for observer absolute orbit estimation.
\item Single observer, fully autonomous: initial target relative orbit estimates are generated on board by IMP + BOD and refined on board by IMP + SOD. GPS is used for observer absolute orbit estimation.
\item Multi-observer, partially autonomous: multiple observers share measurements over the ISL to estimate target relative orbits with improved accuracy. GPS is used for observer absolute orbit estimation.
\item Multi-observer, fully autonomous: multiple observers share measurements over the ISL to estimate target relative orbits with improved accuracy. Inter-satellite bearing angles are used for observer absolute orbit estimation.
\end{enumerate}
With reference to potential applications, Modes 1 and 3 are most applicable to cooperative swarm scenarios in Earth orbit; Mode 2 is most useful for SSA in which unknown targets must be detected and tracked; and Mode 4 is most useful for deep space applications in which GPS is unavailable.

All four navigation modes were investigated in flight.
Fourteen experiment blocks were originally planned\cite{kruger_starfox_2023}, differentiated by swarm geometry, visible targets, presence of maneuvers, presence of the crosslink, presence of GPS measurements, and presence of a-priori relative orbit knowledge.
In total, eighteen StarFOX experiment blocks were executed, over a total of 34 experiment days.
Experiment complexity increased throughout the schedule, with early experiments focusing on single-observer ground-assisted navigation and later experiments focusing on multi-observer autonomous navigation.
Each swarm member could also be commanded independently, to allow multiple simultaneous experiments setups to be executed.

\subsection*{Experiment Execution}

The typical sequence for a StarFOX experiment is as follows.
Initial preparation involves the fusion of data from three sources: NASA Flight Dynamics (FD), NASA mission operators, and Stanford SLAB. 
FD provides predicted swarm ephemerides and planned maneuvers, which are used to provide each spacecraft with its own absolute orbit estimate for initializing navigation, and optionally, a-priori relative orbit estimates for its targets.
SLAB provides a telecommand (TC) table for ARTMS which governs its behavior, e.g. activation of individual ARTMS algorithms, multi-observer measurement use, and granular tuning of IMP, BOD and SOD.
A NASA operator creates command sequences for the spacecraft bus to govern its behavior with respect to StarFOX, including the attitude profile, imaging cadence, crosslink radio activation, ground station contact activities, and so on.
These commands are scheduled in an Absolute Time Sequence plan, which is combined with the TC table into a final product.
Subsequently, NASA's Mission Operations Systems (MOS) and Mission Operations Center perform command validation, using a hardware-in-the-loop simulation environment.
Validated products are uplinked to the swarm.

Payload operations on board the spacecraft consist of several tasks.
First, the spacecraft slews to a pre-defined attitude to ensure consistent swarm visibility, i.e. by aligning its star tracker with the (anti)-velocity direction.
The TC table is unpacked and verified and regular experiment operations commence, including taking and copying star tracker images, running ARTMS applications, and saving ARTMS outputs as telemetry (TM).
Nominal sample times for StarFOX are 60 seconds for IMP, 90 minutes for BOD, and 60 seconds for SOD.
Note that StarFOX operations may be interrupted by various tasks onboard, such as swarm maneuvers and ground passes (which affect satellite attitude) or Globalstar signalling (which reduces GPS accuracy).
Experiments continue for $\sim$24 hours until a daily payload reset.

On-board TM is downlinked during regular ground passes and is provided to NASA Ground Data Systems.
Raw TM from the StarFOX payload and satellite bus is made immediately available to MOS, and is processed to generate reports for payload experiment teams.

\section*{FLIGHT RESULTS}
\label{sec:results}

Flight results for StarFOX are generated in two ways.
In the first case, ARTMS produces telemetry while operating in orbit, and telemetry is downlinked and visualized on the ground.
In the second case, ARTMS inputs (i.e., star tracker images and metadata, GPS data, timing data, and ISL data) are downlinked, and are processed in a flight-like manner using ARTMS flight software running on the ground.
The second case is a ``digital twin'' of the swarm which aims to operate as similarly as possible to ARTMS in flight.
The digital twin was also used for diagnosing and troubleshooting ARTMS during the mission.

Navigation performance is assessed by comparison to ``ground truth'' orbit states, produced by NASA FD via batch processing of GPS flight data.
Expected position errors and uncertainties in the ground truth are significantly less than 10 meters (1$\sigma$).
High-level success criteria for StarFOX are summarized in Table \ref{tab:status}.
Subsequent sections discuss flight results from four experiments in detail, corresponding to the four navigation modes.
\begin{center}
\fontsize{8}{7.2}\selectfont
{\tabulinesep=1.1mm
\captionof{table}{StarFOX mission success status.}
\label{tab:status}
\begin{tabu}{|p{1.5cm}|p{5.5cm}|}
    \hline
    Mission Requirement & Starling shall conduct inter-satellite relative position estimation.\\ \hline
    Min. Success Criteria & Conduct ground-based relative position estimation using onboard measurements.\\ \hline
    Full Success Criteria & Conduct onboard relative position estimation using onboard measurements.\\ \hline
    Completion Status & Full success: onboard images used to produce onboard state estimates for multiple targets.\\
    \hline
\end{tabu}}
\end{center}

\subsection*{Relative Navigation: Single-Observer Multi-Target}

\begin{figure*}[th]
\centering
\includegraphics[width=0.95\textwidth]{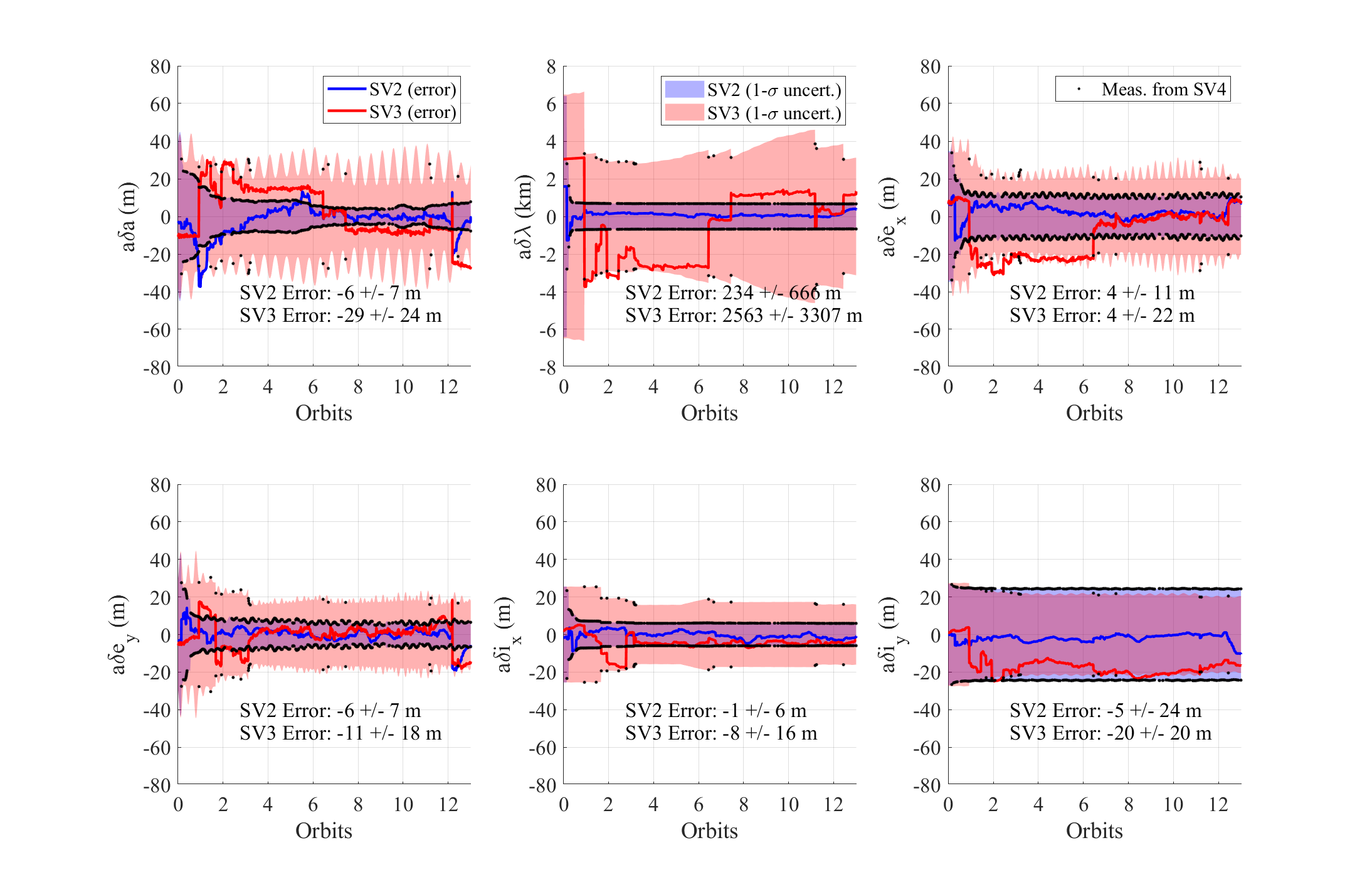}
\caption{ROE state estimation errors and uncertainties (1$\sigma$) for the single-observer multi-target case.}
\label{fig:result1}
\end{figure*}

Figure \ref{fig:result1} presents in-flight estimation performance from SV4 telemetry on 05/16/2024.
In this scenario, SV4 points its star tracker in the anti-velocity direction to image SV2 (at $\sim$50 km range) and SV3 (at $\sim$130 km range).
Relative orbit estimates for SV2 and SV3 are initialized using stale information from the ground (approximately 3-4 days old).
GPS is used to maintain SV4's absolute orbit estimate. 
Relative orbit estimation is performed using bearing angles from SV4 only.
Images were taken every 60 seconds.

It can be seen in Figure \ref{fig:result1} that SV4 successfully tracks and performs orbit estimation for multiple targets simultaneously, which is the first time this capability has been demonstrated in flight.
It is also able to maintain a converged relative orbit estimate over a 21-hour period without requiring target maneuvers, which is another advancement compared to prior flight missions.

Performance displays expected trends for angles-only navigation.
Error is primarily captured by the $\delta \lambda$ ROE, which corresponds to the weakly-observable target range.
Error is also somewhat proportional to range, because at longer distances, the same sensor noise floor corresponds to a larger geometric uncertainty.
Uncertainties (1$\sigma$) for $a \delta \lambda$ at the end of the experiment period are 1.3\% for SV2 and 2.6\% for SV3, as a percentage of target range.
Uncertainties in other ROE are less than 0.02\% of target range because these state components are more strongly observable.
Estimation errors generally remain within 1$\sigma$ bounds, indicating good health of the SOD filter.
In this experiment, the filter assumes bearing angle measurement noise of $40''$ (1$\sigma$); a suggested smaller value of $20''$\cite{palo_agile_2013, kruger_starfox_2023} was found to be insufficiently robust.

Results are also affected by unexpectedly poor target visibility, most obviously for SV3.
In Figure \ref{fig:result1}, black dots indicate a detection of SV3 by SV4's star tracker, and fewer than 20 SV3 measurements were obtained over 12 orbits (whereas measurements SV2 were much more consistent).
This is because the visual magnitude of the swarm was significantly dimmer than expected in orbit.
Furthermore, the image signal-to-noise ratio was lower than expected (e.g., due to the presence of hot pixels).
As a result, targets at distances of more than 100 km were frequently lost in the image and could not be detected by IMP.
Figure \ref{fig:grid} illustrates this phenomenon: SV3 is consistently visible above background noise, whereas the more distant SV2 is not.
Nevertheless, ARTMS was still able to leverage sparse measurements to maintain a reasonable orbit estimate for distant targets.
This is evidence of the overall robustness of the architecture and its ability to maximize angles-only observability.

\begin{Figure}
\centering
\includegraphics[width=0.95\columnwidth]{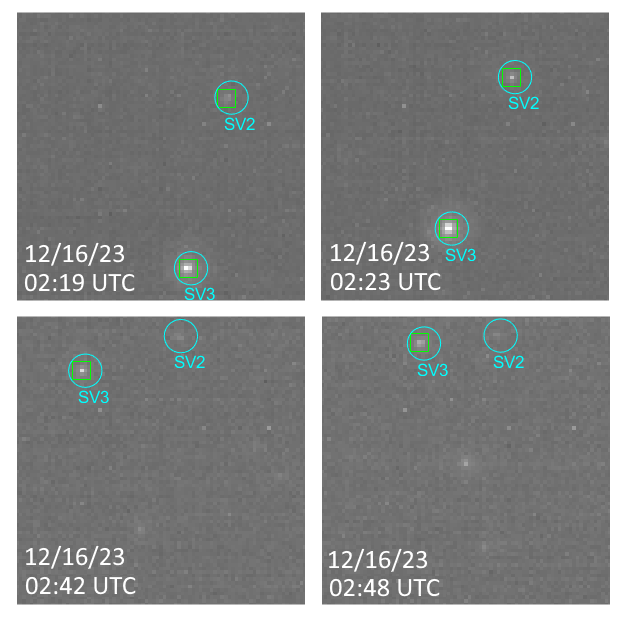}
\captionof{figure}{Target visibility in flight images from SV4. Green boxes indicate successful detection by IMP.}
\label{fig:grid}
\end{Figure}

\subsection*{Relative Navigation: Multi-Observer Multi-Target}

\begin{figure*}[hbt]
\centering
\includegraphics[width=0.95\textwidth]{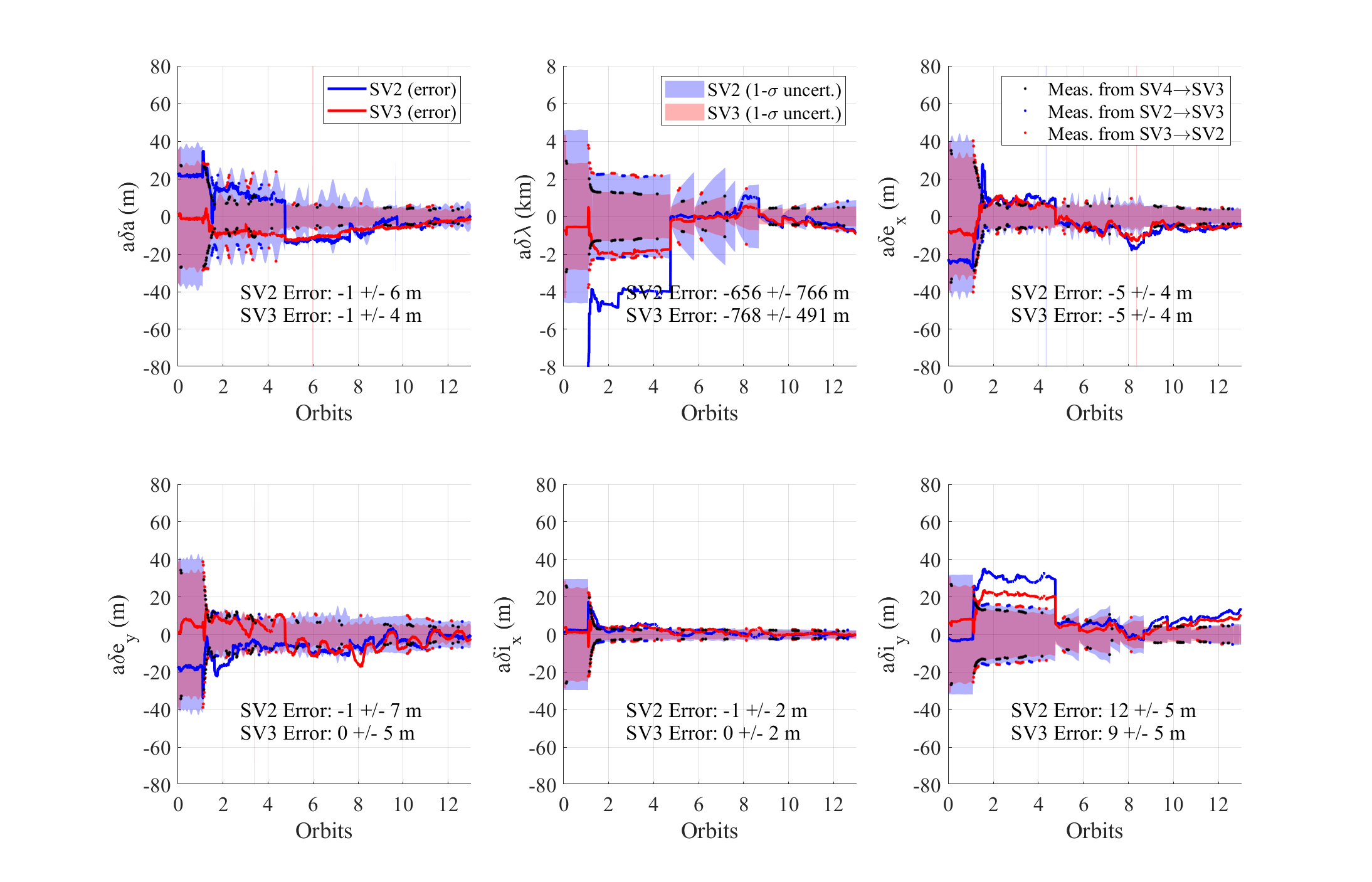}
\caption{ROE state estimation errors and uncertainties (1$\sigma$) for the multi-observer multi-target case.}
\label{fig:result2}
\end{figure*}

Figure \ref{fig:result2} presents in-flight estimation performance from SV4 telemetry on 02/05/2024.
In this scenario, SV4 points its star tracker in the anti-velocity direction to image SV2 (at $\sim$130 km range) and SV3 (at $\sim$80 km range).
Relative orbit estimates for SV2 and SV3 are initialized using stale information from the ground (approximately 3-4 days old).
GPS is used to maintain SV4's absolute orbit estimate.
Relative orbit estimation is performed using bearing angles from multiple observers over the ISL, such that SV4 additionally receives crosslink measurements from SV2 (measuring SV3 and SV4) and SV3 (measuring SV2).
Images were taken every 120 seconds.

The colored dots in Figure \ref{fig:result2} mark instances when a measurement broadcast by a remote observer was used to update local state estimates on SV4.
It can be seen that SV4 utilizes angles-only measurements from all three swarm spacecraft, which is the first time this capability has been demonstrated in flight.

The results also reveal several advantages of distributed multi-observer navigation.
Firstly, SV4 only obtains 5 local measurements of SV2 throughout the entire period.
It must instead rely on measurements of SV2 from SV3 (and/or SV3 from SV2) to update its on-board orbit estimate for SV2.
Thus, multiple observers are able to provide added robustness when measurements are sparse.
Secondly, state uncertainty is reduced compared to an equivalent single observer case.
Uncertainties (1$\sigma$) for $a \delta \lambda$ at the end of the experiment period are 0.6\% for SV2 and 0.6\% for SV3, as a percentage of target range.
Uncertainties in other ROE are less than 0.01\% of target range.
The additional geometric information provided by multi-observer measurements significantly improves angles-only observability.

Several anomalies are also evident in Figure \ref{fig:result2}, such as sudden decreases in the $a\delta\lambda$ uncertainty at several time instants (e.g. after approximately 6 orbits).
This is due to unintended behavior of ARTMS when the SOD filter resets.
As part of the ISL message, swarm members broadcast their absolute orbit estimate, which allows the receiving spacecraft to match any remote measurements to local state estimates.
Typically, this absolute orbit estimate is taken from the on-board SOD filter estimate.
However, when SOD re-initializes (e.g. due to a large number of skipped measurement updates), the SOD estimate becomes temporarily invalid, and the on-board GNSS solution is sent instead as backup.
When SV4 received this GNSS solution, it interpreted it as a measurement that could be used to update the relevant target’s state estimate.
GNSS provides improved observability in this scenario which led to a sudden improvement in that target's state uncertainty.
(Remote GNSS navigation updates can be disabled within ARTMS but were left active during the 02/05/24 experiment.)

Earlier in the experiment, rapid covariance growth is observed after these instances, as relative orbit uncertainties stabilize after a GNSS update.
Towards the end of the experiment, however, the steady-state uncertainty provided by multi-observer bearing angles is more consistently maintained. 
This implies the noise floor of multi-observer navigation is indeed at a similar level to the figure, even if the irregular GNSS updates may have sped up convergence.

Finally, note that there are certain periods when state estimation error ventures outside 1$\sigma$ bounds, especially for $a\delta\lambda$ and $a\delta i_y$.
However, errors do remain within $3\sigma$ bounds.
This phenomena is believed to be caused by drifts in measurement time-tags between swarm members, discussed further in the following section.

\subsection*{Relative Navigation: Autonomous Initialization}

\begin{figure*}[hbt]
\centering
\includegraphics[width=\textwidth]{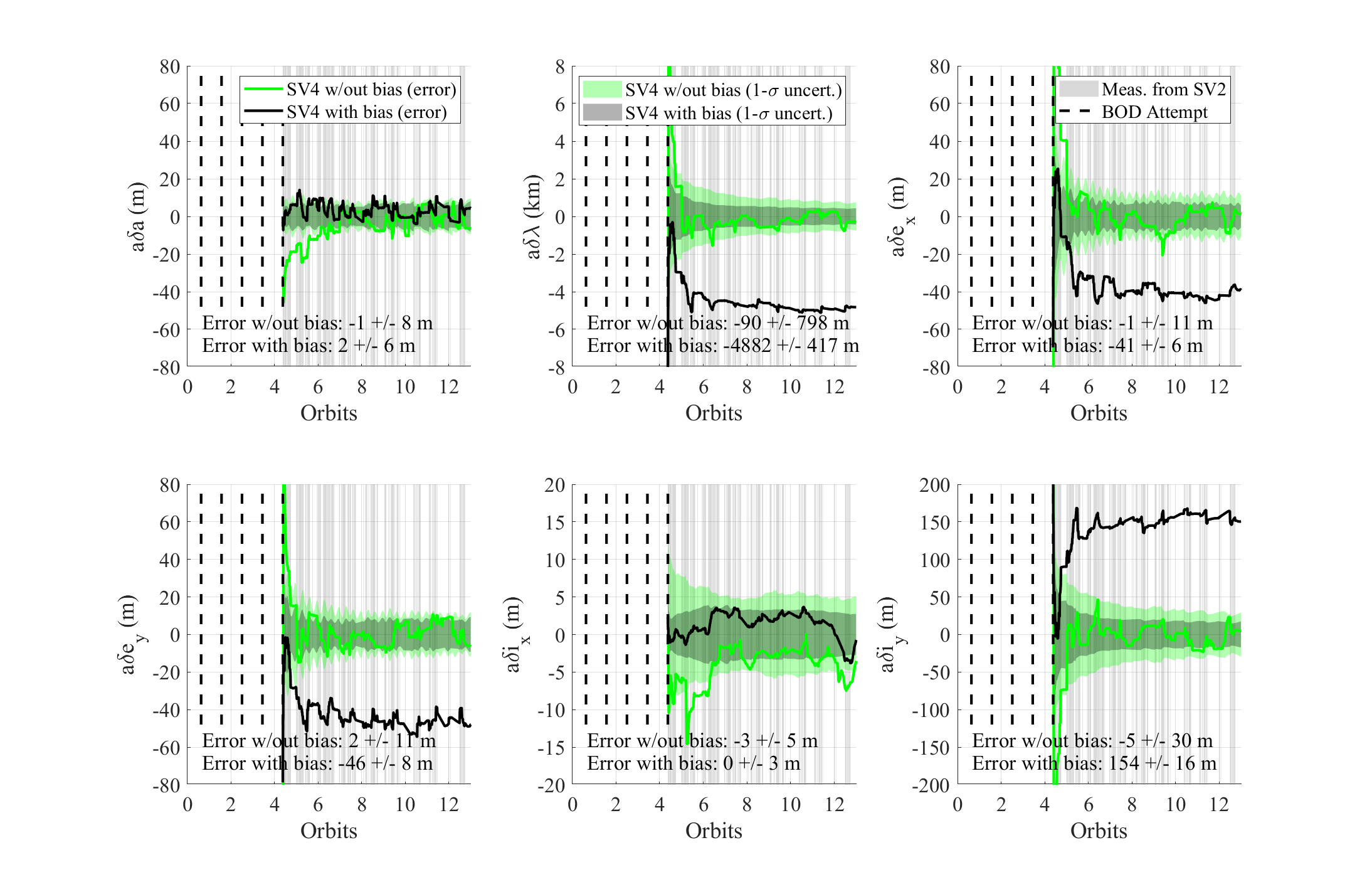}
\caption{ROE state estimation errors and uncertainties (1$\sigma$) for the autonomous initialization case.}
\label{fig:result3}
\end{figure*}

Figure \ref{fig:result3} presents in-flight estimation performance from SV2 telemetry on 05/08/2024.
In this scenario, SV2 points its star tracker in the velocity direction to image SV4 (at $\sim$50 km range).
No a-priori relative orbit information for SV4 is provided.
GPS is used to maintain SV4's absolute orbit estimate.
SV2 begins by performing multi-hypothesis tracking with IMP to detect potential targets in the FOV.
Every 1.5 hours, BOD attempts to compute a relative orbit initialization using IMP measurement batches.
If the computed orbit is reasonable (i.e., is within state and uncertainty tolerances), it is used to initialize SOD and formal navigation commences.
Images were taken every 60 seconds.

Figure \ref{fig:roi} presents overlaid regions of interest from one orbit of images, as recorded in IMP telemetry.
The initial ROE estimate was produced using a batch of 36 measurements, corresponding to the upper half of the ellipse in the figure.
Figure \ref{fig:result4} presents the measurement residuals for $\delta \lambda$ state samples and the output $\delta \lambda$ estimate, compared to the true ISD from FD.
Although the initial BOD range estimate possesses errors of 8 km (approximately 15\% of the true range), this initial error is well-captured by the onboard uncertainty estimate.
The convex form of the residual curve in Figure \ref{fig:result4} also matches pre-flight expectations, as per Figure \ref{fig:BOD}.
Note that some difficulties were created by inconsistent image availability; of the 415 images expected to be available to IMP during this period, only 228 were successfully provided to ARTMS by the satellite bus.
Despite this, IMP was able to track the unknown target amid the background starfield and other transient RSO, and BOD was able to produce a reasonable initialization.
This is the first time this capability has been demonstrated in flight.

\begin{Figure}
\centering
\includegraphics[width=\columnwidth]{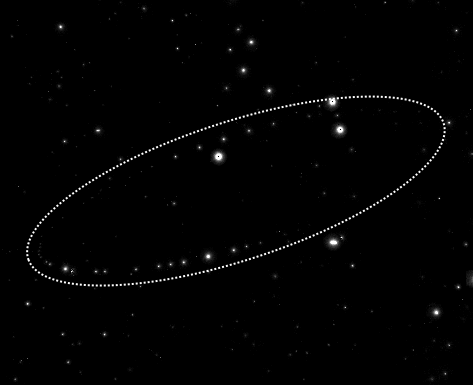}
\captionof{figure}{Overlaid regions of interest from one orbit of images, for SV2 viewing SV4 in the PSE formation. The point sources corresponding to the SV4's relative orbit are bordered by the ellipse.}
\label{fig:roi}
\end{Figure}

\begin{Figure}
\centering
\includegraphics[width=0.95\columnwidth]{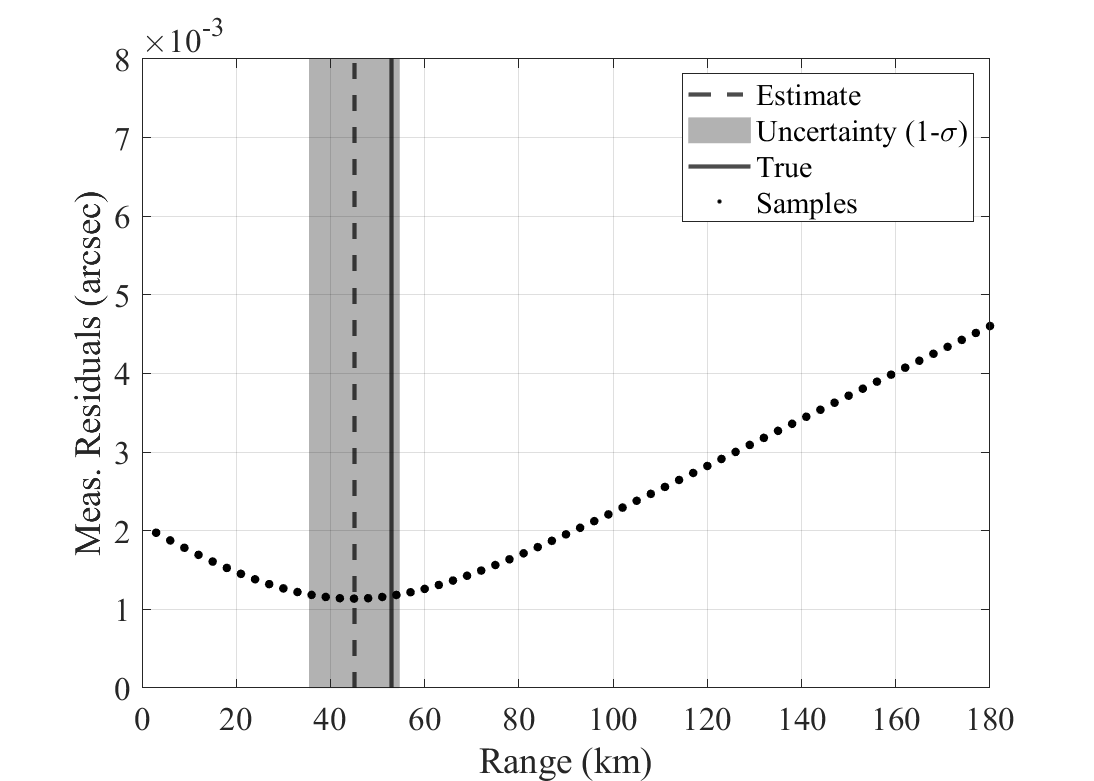}
\captionof{figure}{Initial target range estimation errors and uncertainties from BOD.}
\label{fig:result4}
\end{Figure}

Figure \ref{fig:result3} presents subsequent estimation performance.
IMP + BOD autonomously initializes navigation on the fifth attempt, after approximately 4 orbits.
The black line corresponds to estimation errors from in-flight telemetry, where it is clear that even if state uncertainties are converging, a strong bias is present in the state error for $\delta \lambda$, $\delta e_x$, $\delta e_y$, and $\delta i_y$.
It was determined that this was produced by timing errors on board the satellite.
The timestamps associated with star tracker images and metadata appeared to drift over time, and became unsynchronized with the primary clock (which was synchronized to GPS time).
The green line in Figure \ref{fig:result3} corresponds to estimation performance after the image timing errors are corrected, which removes the bias in all estimated ROE.
The applied time correction for this experiment was -0.4 seconds, and was found to vary between spacecraft.

\subsection*{Absolute Navigation: Multi-Observer Multi-Target}

\begin{figure*}[hbt]
\centering
\includegraphics[width=0.95\textwidth]{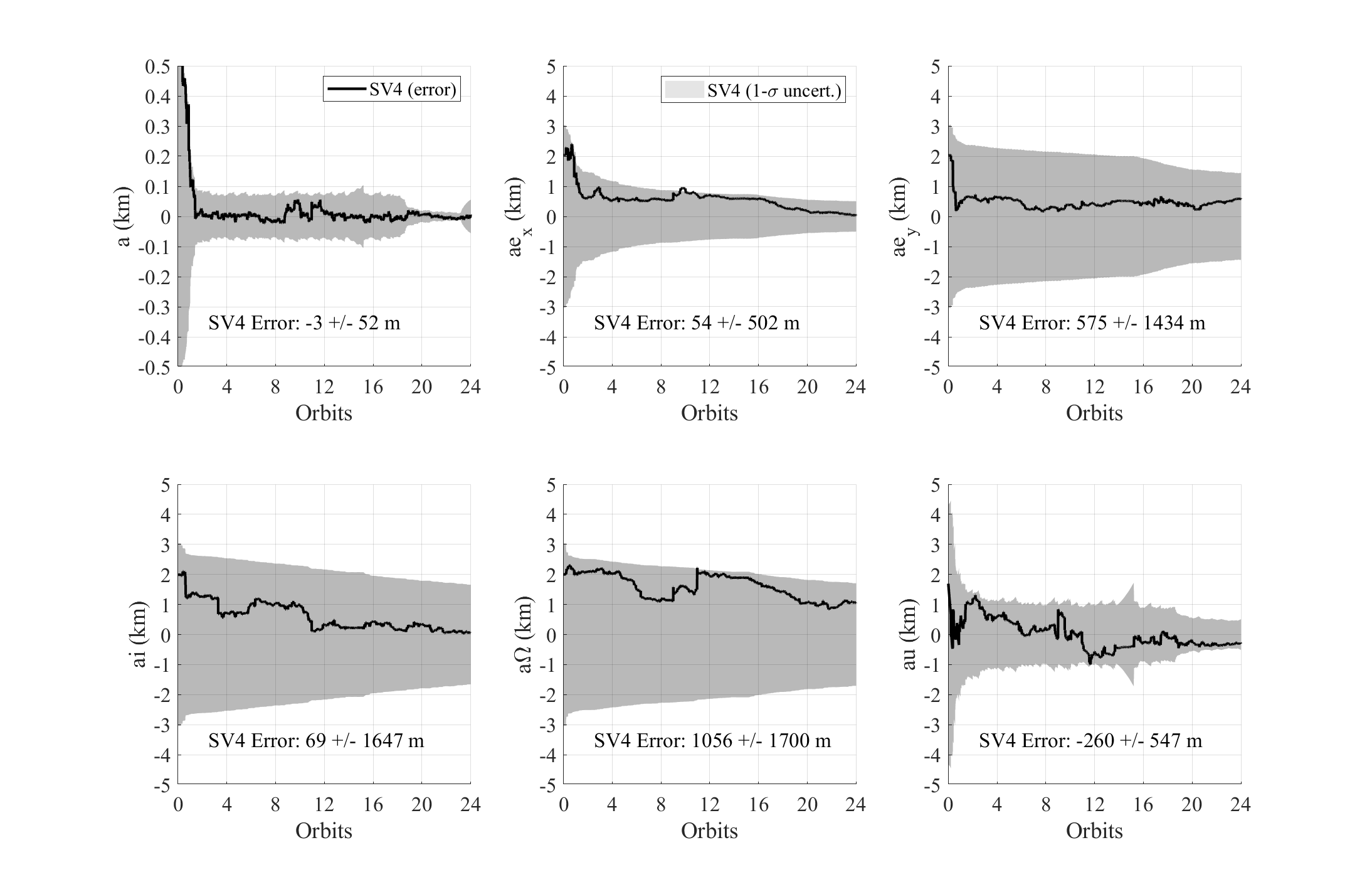}
\caption{OE state estimation errors and uncertainties (1$\sigma$) for the multi-observer case.}
\label{fig:result5}
\end{figure*}

Demonstrations of Navigation Mode 4 were attempted on 05/09/24 and 05/16/24 in flight but crosslink issues prevented successful execution. 
It was later demonstrated using flight data from 03/24/24, post-processed by ARTMS flight software running on the ground.
The following scenario applies three ARTMS instances (emulating SV2, SV3 and SV4) to process approximately 3300 star tracker images over 40 hours, with simulated ISL transmissions between all members.
On-board state estimates are initialized using artificially degraded orbit knowledge, to emulate a coarse initialization with uncertainties on the scale of several kilometers.
After initialization, no additional external absolute orbit knowledge is provided (i.e. no GPS data is processed).
Both absolute and relative orbit determination are performed using bearing angles only.

Figure \ref{fig:result5} presents absolute orbit estimation performance for SV4.
Orbit element errors are normalized by the semi-major axis for geometric interpretation.
Absolute orbit estimation using only inter-satellite bearing angles is particularly challenging due to the weak observability of the problem.
However, if multiple observers are present and are obtaining measurements of a common target (and each other), enough geometric information is present to facilitate long-term absolute orbit convergence without external updates \cite{hu_angles_2021, kruger_observability_2022, kruger_starfox_2023}.

Here, SV4 observes SV2 and SV3 as targets, but also receives crosslink measurements from SV3 (observing SV2 and SV4) and SV2 (observing SV3). 
The coarse state initialization is successfully refined over the experiment period by leveraging multi-observer images.
In the RTN frame of SV4, this corresponds to an initial position error of 6950 m which is reduced to a final error of 1200 m.
If initial errors were halved, final position errors were 850 m.
Convergence of the state estimate is slow, due to challenging observability, but estimation errors remain within 1$\sigma$ uncertainty bounds, indicating reasonable filter health.
This is the first time this capability has been demonstrated using flight software and flight data.

Figure \ref{fig:result6} displays the ISL measurements received by SV4. 
The most consistently visible link was SV4 observing SV2, whereas other observations were more sparse (due to the visibility issues previously described).
The variation in measurement rates between seemingly equivalent spacecraft pairs (e.g. SV4 $\rightarrow$ SV3 and SV3 $\rightarrow$ SV4) illustrates an additional difficulty of applying optical measurements: visibility varies with specific orbit geometry and attitude, and image properties vary with specific VBS hardware.
Scenario-specific tuning of image processing algorithms is therefore beneficial.

\begin{Figure}
\centering
\includegraphics[width=\columnwidth]{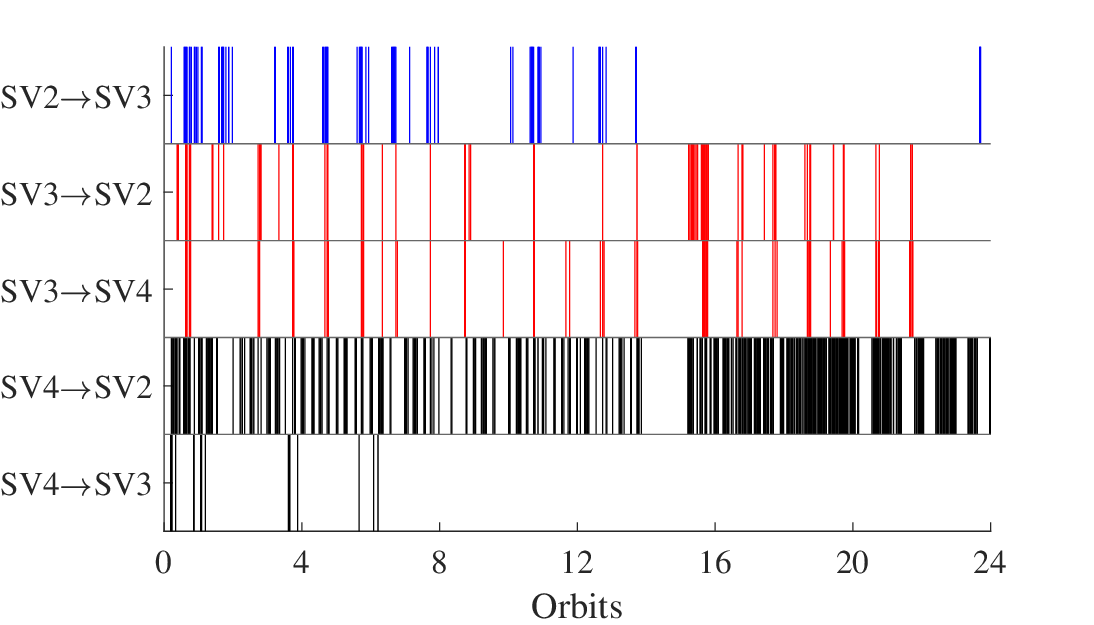}
\captionof{figure}{Measurements received by SV4. A bar indicates a measurement was received at that epoch.}
\label{fig:result6}
\end{Figure}

\subsection*{Performance Summary}

A series of experiment goals was defined for StarFOX prior to launch, informed by overall Starling objectives and performance during pre-flight simulations and testing \cite{kruger_starfox_2023}.
These goals, and StarFOX outcomes during the initial experiment phase, are presented in Table \ref{tab:result}.
All thresholds were successfully achieved and the majority of goals were successfully achieved, through either in-flight experiments or on-ground flight data processing.

Objectives 6 and 7, which relate to swarm maneuvers, have not yet been discussed.
Station-keeping maneuvers were executed during StarFOX experiments on 05/09/24 and 05/16/24, with the objective of observing a corresponding decrease in state uncertainty.
Known maneuvers are expected to improve angles-only observability by aiding the filter in disambiguating target range \cite{woffinden_observability_2009, damico_argon_2013}.
Although maneuvers were successfully tracked on board, the executed delta-vs were small ($\sim$0.01 m/s) and did not significantly improve state uncertainties as a result.

\begin{table*}[htb]
\begin{center}
\fontsize{8}{7.2}\selectfont
{\tabulinesep=1.1mm
\captionof{table}{Summary of StarFOX performance goals and initial flight results.}
\label{tab:result}
\begin{tabu}{|p{0.5cm}|p{4.5cm}|p{2.7cm}|p{2.0cm}|p{4.5cm}|}
    \hline
    \# & Objective & Threshold & Goal & Initial Result \\
    \hline
    1 & Tracking multiple targets at one time & 1 target & 3 targets & 2 targets (in flight) \\
    2 & Accuracy of relative position knowledge with one observer & Estimate produced & 1\% error relative to range & 0.5\% error relative to range (in flight)\\
    3 & Convergence time for relative position knowledge with one observer & Convergence observed & 6 orbits & 2 orbits (in flight)\\
    4 & Accuracy of relative position knowledge with multiple observers & Estimate produced & 0.1\% error relative to range & 0.5\% (in flight); 0.1\% (on ground) \\
    5 & Convergence time for relative position knowledge with multiple observers & Convergence observed & 2 orbits & 1 orbit (in flight)\\
    6 & Accuracy of relative position knowledge in presence of maneuvers & Estimate produced & 0.1\% error relative to range & Estimate produced (in flight)\\
    7 & Convergence time for relative position knowledge in presence of maneuvers & Convergence observed & 2 orbits & Convergence observed (in flight)\\
    8 & Accuracy of absolute position knowledge with multiple observers & Estimate produced & 1 km error & 850 m error (on ground) \\
    9 & Convergence time for absolute position knowledge with multiple observers & Convergence observed & 24 hours & Convergence observed (on ground)\\
    10 & Time to produce autonomous relative orbit initialization & Initialization produced & 2 orbits & 5 orbits (in flight)\\
    \hline
\end{tabu}}
\end{center}
\end{table*}

\subsection*{Lessons Learned}

The flight experiences of StarFOX give rise to a number of lessons learned, which may be applied to any distributed mission employing vision-based angles-only navigation.
The most critical of these is the overwhelming importance of realistic pre-flight simulation and validation, in software and hardware.

During validation of spacecraft flight software, it is typical to conduct simulations which attempt to recreate the orbital environment as closely as possible.
For StarFOX, the most crucial aspect is generation of realistic space images; significant attention was paid to producing appropriate imagery via an Optical Stimulator testbed with hardware-in-the-loop elements \cite{beierle_os_2019, kruger_starfox_2023}.
Efforts were made to incorporate non-ideal measurement properties in regards to image noise, camera calibration errors, and measurement sparsity.
Even so, in-flight conditions proved significantly more challenging than any assumptions made before launch. 
Target spacecraft were much dimmer than expected, sun intrusion into the FOV was not effectively modeled, and the image signal-to-noise ratio was too optimistic. 
In response, in-flight software updates were needed to enhance the adaptability of ARTMS to different optical conditions, via improved noise filtering and image normalization.
This greatly improved performance in the second half of the mission.
More realistic simulations may also have justified higher-level changes to the experiment design.
For example, more accurate determination of target visual magnitude could have supported selection of a smaller nominal ISD. 

With respect to validation on spacecraft hardware, attention should be devoted to providing valid sensor inputs during integration and testing (I\&T).
Unfortunately, resource constraints meant that image measurements could not be provided to ARTMS during integration with the satellite bus, and thus, the majority of execution paths were not exercised.
As a result, ARTMS encountered software crashes during its first experiments, caused by surpassing onboard memory limits.
Though memory profiling had been performed pre-launch, the lack of realistic images during profiling meant that results were not representative.
Second, as discussed, there was a variable drift between star tracker time-tags and the on-board clock aboard all swarm spacecraft.
This could potentially have been discovered earlier if simultaneous image and GPS data was available during I\&T.
Instead, an in-flight software update was needed to allow user-specified time offsets for incoming star tracker data.
It is also possible to estimate differential clock offsets between observers online using only inter-satellite angles \cite{kruger_artms_2021} which will be explored during StarFOX+.
More generally, this displays the importance of accurate time synchronization for navigation (especially if navigation is to be performed autonomously, e.g. without regular GPS time updates).
The resulting state errors were significant and sources of time for different measurement types must be well-understood. 
%

A final note is the importance of flexibility and robustness to overall system design. Efforts were made during the development of ARTMS to allow the system to adapt to differing conditions, in either an autonomous fashion (via different software execution paths, contingencies, and checks) or in a user-defined fashion (via tunable telecommand parameters). These decisions are what have enabled successful navigation even when in-flight conditions have proven particularly challenging.

\subsection*{Future Work}

The Starling primary mission concluded in May 2024.
However, StarFOX efforts will continue as part of Starling 1.5, a mission extension implemented via ground and onboard software updates.
The capabilities of ARTMS will be augmented to include: 1) improved detection and tracking of unknown resident space objects for SSA, 2) faster orbit initialization for quickly evolving scenarios such as space rendezvous, 3) online estimation of auxiliary state components such as clock errors, drag coefficients and solar radiation pressure coefficients, 4) data fusion with inter-satellite crosslink ranging to achieve higher estimation accuracy, and 5) a fly-by inspection of a target satellite with control in the loop. 
\section*{CONCLUSION}
\label{sec:conclusion}

This paper presents an overview of the Starling Formation-flying Optical eXperiment (StarFOX), which is the first in-flight demonstration of autonomous angles-only navigation for a spacecraft swarm.
StarFOX is a core experiment payload of the NASA Starling mission, managed by the NASA Ames Research Center.
Starling consists of four identical 6U CubeSats and was launched in July 2023.
Prior demonstrations of angles-only navigation in orbit have performed navigation for single observers and single targets only, and have required external orbit information and/or system maneuvers to achieve orbit determination.
To overcome these limitations, StarFOX applies the new angles-only Absolute and Relative Trajectory Measurement System (ARTMS).
The ARTMS architecture  provides autonomous, distributed, and scalable navigation for distributed systems in deep space using inexpensive optical sensors.
Its multi-observer multi-target framework is able to achieve complete swarm orbit determination with minimal external orbit information and without maneuvers.

ARTMS consists of three modules: IMage Processing (IMP), Batch Orbit Determination (BOD), and Sequential Orbit Determination (SOD).
The IMP module provides batches of measurements to each observed target using time-tagged images from the onboard camera and a single coarse estimate of the observer's orbit.
The BOD module uses these batches of bearing angles to each target, along with the observer orbit estimate, to provide estimates of the orbits of each observed system member.
Finally, the SOD module refines system state estimates by seamlessly fusing measurements from multiple observers received over an inter-satellite crosslink within an unscented Kalman filter.

Initial StarFOX experiments have applied ARTMS in flight to successfully perform distributed swarm navigation.
Star trackers are used to image other swarm spacecraft and obtain bearing angle measurements, which are optionally shared over the crosslink.
Observers are able to navigate for multiple target spacecraft and display maneuver-free angles-only convergence, achieving relative position estimation uncertainties of 1.3\% of target range (1$\sigma$).
Multiple observers are able to cooperate to estimate relative orbits with an improved 0.6\% position uncertainty relative to range (1$\sigma$).
StarFOX has also provided the first demonstrations of autonomous initialization of relative navigation, in which an unknown target is detected and tracked by an observer and a relative state initialization is computed in flight.
Finally, absolute navigation using only inter-satellite bearing angles was investigated, whereupon coarse initial orbit errors were refined from several kilometers to less than 1 km.

StarFOX encountered several challenges in flight, including poorer than expected target visibility and time synchronization issues, both of which degraded navigation performance.
It is likely these could have been mitigated pre-launch with improved image simulation capabilities and more extensive hardware-in-the-loop testing.
Nevertheless, StarFOX has achieved the majority of its pre-flight goals during its initial experiment period.
It therefore presents a significant step towards the autonomy of future distributed missions and more widespread application of vision-based navigation in space.

\subsection*{Acknowledgments}

NASA Starling is funded by the Small Spacecraft Technology Program (SSTP) within the NASA Space Technology Mission Directorate. StarFOX development was supported by SSTP cooperative agreement number 80NSSC18M0058 and by the Air Force Office of Scientific Research award number FA9550-21-1-0414.

The authors acknowledge the invaluable contributions of Joshua Sullivan and Adam W. Koenig towards the conception and development of ARTMS and StarFOX.
They also thank the NASA Ames Starling team for their tireless support of StarFOX efforts.

\bibliography{bib.bib}
\bibliographystyle{unsrt}
\end{multicols*}

\end{document}